\documentclass[10pt,twocolumn,letterpaper]{article}

\usepackage{cvpr}              

\newcommand\mypara[1]{\vspace{0.6mm}\noindent\textbf{#1}}

\usepackage{graphicx}
\usepackage{amsmath}
\usepackage{amssymb}
\usepackage{gensymb}
\usepackage{booktabs}
\usepackage{adjustbox}
\usepackage{enumitem}
\usepackage{multirow}
\usepackage{array}
\usepackage{algorithm}
\usepackage{algpseudocode}
\algdef{SE}[DOWHILE]{Do}{doWhile}{\algorithmicdo}[1]{\algorithmicwhile\ #1}%
\usepackage[pagebackref,breaklinks,colorlinks]{hyperref}

\usepackage[capitalize]{cleveref}
\crefname{section}{Sec.}{Secs.}
\Crefname{section}{Section}{Sections}
\Crefname{table}{Table}{Tables}
\crefname{table}{Tab.}{Tabs.}

\newcommand{\argo}{Argoverse\xspace}

\newcommand{\kitti}{KITTI\xspace}
\usepackage{amssymb}
\usepackage{amsmath,amsfonts,amssymb}
\usepackage{amsopn}
\usepackage{bm} 
\usepackage{multirow}

\newcommand{\ProbOpr}[1]{\mathbb{#1}}

\newcommand{\var}[2]{%
\ifthenelse{\equal{#2}{}}{\ProbOpr{VAR}_{#1}}
{\ifthenelse{\equal{#1}{}}{\ProbOpr{VAR}\left[#2\right]}{\ProbOpr{VAR}_{#1}\left[#2\right]}}} 

\newcommand{\eat}[1]{}

\newcommand{\APBEV}{AP$_\text{BEV}$\xspace}
\newcommand{\AP}{AP$_\text{3D}$\xspace}

\DeclareRobustCommand{\eg}{e.g.\@\xspace}
\DeclareRobustCommand{\ie}{i.e.\@\xspace}

\def\cvprPaperID{11670} 
\def\confName{CVPR}
\def\confYear{2022}

\usepackage{overpic}
\usepackage{enumitem} 
\usepackage{overpic} 
\usepackage{color}

\definecolor{turquoise}{cmyk}{0.65,0,0.1,0.3}
\definecolor{purple}{rgb}{0.65,0,0.65}
\definecolor{dark_green}{rgb}{0, 0.5, 0}
\definecolor{orange}{rgb}{0.8, 0.6, 0.2}
\definecolor{red}{rgb}{0.8, 0.2, 0.2}
\definecolor{darkred}{rgb}{0.6, 0.1, 0.05}
\definecolor{blueish}{rgb}{0.0, 0.3, .6}
\definecolor{light_gray}{rgb}{0.7, 0.7, .7}
\definecolor{pink}{rgb}{1, 0, 1}
\definecolor{greyblue}{rgb}{0.25, 0.25, 1}

\usepackage{blindtext}

\renewcommand{\paragraph}[1]{\vspace{1em}\noindent\textbf{#1}.}

\makeatletter
\newcommand*{\@rowstyle}{}
\newcommand*{\rowstyle}[1]{
 \gdef\@rowstyle{#1}%
 \@rowstyle\ignorespaces%
}
\newcolumntype{=}{
>{\gdef\@rowstyle{}}%
}
\newcolumntype{+}{
>{\@rowstyle}%
}
\newcommand{\PreserveBackslash}[1]{\let\temp=\\#1\let\\=\temp}
\newcolumntype{C}[1]{>{\PreserveBackslash\centering}p{#1}}
\makeatother

\begin{document}
\title{Ithaca365: Dataset and Driving Perception \\under Repeated and Challenging Weather Conditions}

\author{
Carlos A. Diaz-Ruiz$^{1}$\hspace{10pt}
Youya Xia$^{1}$\hspace{10pt}
Yurong You$^{1}$\hspace{10pt}
Jose Nino$^{1}$\hspace{10pt}
Junan Chen$^{1}$\\
Josephine Monica$^{1}$\hspace{10pt}
Xiangyu Chen$^{1}$\hspace{10pt}
   Katie Luo$^{1}$\hspace{10pt}
   Yan Wang$^{1}$\hspace{10pt}
    Marc Emond$^{1}$\hspace{10pt}\\
  Wei-Lun Chao$^{2}$\hspace{10pt}
   Bharath Hariharan$^{1}$\hspace{10pt}
   Kilian Q. Weinberger$^{1}$\hspace{10pt}
   Mark Campbell$^{1}$\\
   $^1$Cornell University\hspace{14pt}$^2$The Ohio State University

   }
{\tt\small }

\maketitle

\begin{abstract}

Advances in perception for self-driving cars have accelerated in recent years due to the availability of large-scale datasets, typically collected at specific locations and under nice weather conditions. Yet, to achieve the high safety requirement, these perceptual systems must operate robustly under a wide variety of weather conditions including snow and rain.  In this paper, we present a new dataset to enable robust autonomous driving via a novel data collection process --- data is \textbf{repeatedly} recorded along a 15 km route under \textbf{diverse scene} (urban, highway, rural, campus), \textbf{weather} (snow, rain, sun), time (day/night), and \textbf{traffic conditions} (pedestrians, cyclists and cars). The dataset includes images and point clouds from cameras and LiDAR sensors, along with high-precision GPS/INS to establish correspondence across routes. The dataset includes road and object annotations using \textbf{amodal masks} to capture partial occlusions and 3D bounding boxes. We demonstrate the uniqueness of this dataset by analyzing the performance of baselines in amodal segmentation of road and objects, depth estimation, and 3D object detection. The repeated routes opens new research directions in object discovery, continual learning, and anomaly detection. Link to Ithaca365: \small{\url{https://ithaca365.mae.cornell.edu/}} 

\end{abstract}

\section{Introduction}
     The self-driving car research community has made major advancements in computer vision and perception by relying on vast amounts of real world sensory datasets. To date, many datasets have been published~\cite{argoverse,waymo_open_dataset,geiger2012we,nuscenes2019,lyft2019}, including some with benchmark challenges associated with different perception tasks. These tasks include  2D image detection \cite{ren2015faster,lin2017feature}, depth estimation (stereo \cite{wang2018anytime,chang2018pyramid} and monocular \cite{fu2018deep}), 3D object detection (stereo \cite{konigshofrealtime,E2EPL,pseudoLiDAR} and LiDAR \cite{chen2019fast,yang2018pixor,wang2019frustum}), semantic \cite{zhao2017pyramid} and instance segmentation \cite{li2017fully, li2016amodal}. The KITTI dataset \cite{geiger2012we} has been used for many of these benchmarks due to its comprehensive annotations and ground truths. However, like many of the others \cite{Cordts2016Cityscapes}, the KITTI dataset was collected  in only sunny/clear conditions. 
     
     There is a need in the research community for large-scale datasets with  multi-modal sensor data (LiDAR, cameras, and GPS/IMU) in \emph{adverse weather conditions} to be able to train and test the performance and robustness of object detectors, image segmentation, and depth estimation algorithms across varying conditions. The lack of data in these challenging conditions ultimately limits the generalizability of the perception approaches, and as such, limits the applicability of self-driving cars only to environments with nice weather conditions. Recently, a few datasets with more diverse traffic, scenes, and weather conditions have been published \cite{nuscenes2019,argoverse,maddern20171}. These datasets include rain, low light conditions at night, or very bright conditions, but do not include snowy conditions. One such dataset with snowy conditions is the Canadian Adverse Driving Conditions Dataset \cite{DBLP:journals/corr/abs-2001-10117}, a dataset collected in Waterloo. However, this dataset is primarily snowy conditions, and includes just 3D object labels.

\begin{figure*}[t!]
    
  \centering 
  \includegraphics[width=0.95\linewidth]{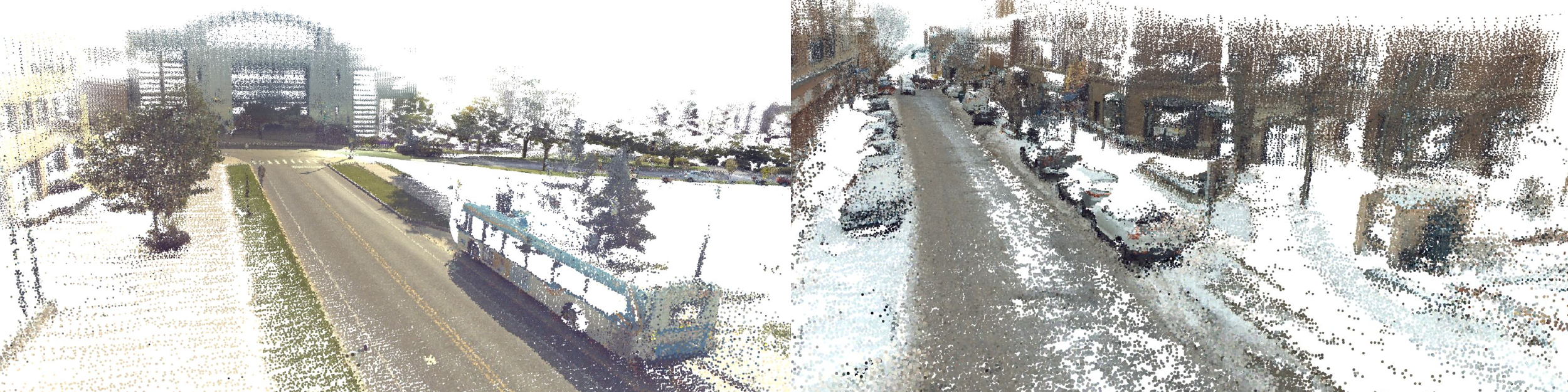}
  \vskip-5pt
  \caption{\small Point clouds generated using calibrated images, LiDAR points, and GPS/IMU data. Left: Point cloud from a sunny day while driving near buildings on campus. Right: Point cloud from a snow day (\ie, white on the road) through an urban portion of the route.}
  \vspace{-10pt}
  \label{pointclouds-fig}
\end{figure*}

Moreover, to our knowledge an amodal segmentation dataset with varying weather conditions does not exist. Amodal perception~\cite{li2016amodal,zhu2017semantic} aims to perceive and understand the physical structure and semantics of occluded objects and scenes. Amodal perception is critical for autonomous driving as the ability to infer the whole shape of objects (\eg, other vehicles, pedestrians, and road) around a mobile platform  allows for safer and more efficient navigation. This is especially useful in highly cluttered environments or complex traffic scenes, where determining  safe, collision free paths based only on visible cues is difficult. The KINS dataset \cite{Qi_2019_CVPR} augments KITTI~\cite{Geiger2013IJRR} with amodal foreground masks and while \cite{lu2020semantic} augments Cityscapes~\cite{Cordts2016Cityscapes} with amodal background classes, but both of these are only in sunny and clear conditions. There are a few recent developments that address amodal segmentation \cite{li2016amodal,lu2020semantic, Qi_2019_CVPR}. Unfortunately, existing literature lacks progress in these areas, primarily due to the lack of large-scale, diverse datasets annotated with amodal masks. 
     
In this paper we provide three major contributions. First, we release a large-scale, \textit{weather and environmental diverse}, \textit{amodal} dataset obtained through \textit{repeatedly} driving a 15km route over a 1.5 year period. The key attributes are:
\begin{itemize} [nosep,topsep=1pt,parsep=1pt,partopsep=1pt, leftmargin=*]
\item \textbf{weather diversity}: snow, rain, sunny, cloudy, night
\item \textbf{environmental diversity}: urban traffic, highway, rural, pedestrian heavy university campus
\item \textbf{amodal}: \!roads, cars, \!pedestrians occluded by snow,objects
\item \textbf{repeated route}: 40 data collections on a 15km route
\end{itemize}
More than 680k frames were collected with LiDAR, images and GPS data. Figure~\ref{pointclouds-fig} shows some 3D visualizations.
Because of the repeatability of the route, amodal labeling of background classes such as road and even parked cars are easily facilitated across the extremely varied conditions without tedious annotation. Second, we suggest new metrics for the proper evaluation of amodal segmentation of 3D objects and background classes such as road. Finally, we  develop baseline model architectures to highlight the utility of the proposed dataset for key self-driving car perception tasks: amodal background segmentation (road), amodal instance segmentation (car, pedestrians, cyclists, motorcyclists, truck, bus, etc.), 3D object detection, depth estimation across varying weather conditions, route types, and leveraging repeatability for unsupervised labeling.
\section{Related Work}
\label{sec:related}
\mypara{Autonomous driving datasets.} 
The \textbf{KITTI} dataset \cite{geiger2012we} is one of the most popular auto driving datasets and has paved the way for the development of deep learning algorithms for 2D and 3D object detection, stereo depth estimation, and segmentation tasks; all data was collected in daylight and nice weather. The \textbf{Oxford RoboCar} dataset \cite{maddern20171} contains stereo cameras, 2D and 3D LiDAR, GPS/IMU data; this dataset capture various weather conditions along the same route but only reports one route with snow conditions.  Additionally, the dataset was released with a focus on localisation and thus does not include 2D/3D object or segmentation annotations. The recent \textbf{GROUNDED} dataset \cite{ort2021radar} is for localization in varying weather conditions, with a focus on ground penetrating radar.  The \textbf{Canadian Adverse Driving Conditions (CADC)} dataset~\cite{DBLP:journals/corr/abs-2001-10117} includes 7K frames with 3D object labels in snow conditions with surround vision, LiDAR data, and ground truth motion. The \textbf{ApolloScape} open dataset \cite{wang2019apolloscape} contains LiDAR, cameras, and GPS in cloudy and rainy conditions, along with brightly sunlit situations. The dataset includes large scale 3D car annotations and segmentation labels. The authors indicate that data in snow conditions will be added, but none has been released. There have been various other large-scale datasets: The \textbf{nuScenes} dataset \cite{nuscenes2019}, \textbf{Argoverse} dataset \cite{argoverse}, \textbf{Waymo} open dataset \cite{waymo_open_dataset}, and \textbf{Lyft Level 5} dataset \cite{lyft2019}. NuScenes has 40k annotated frames, with LiDAR, cameras and radar. Waymo is the largest dataset with 200k frames and 5 LiDARs, and cameras. Lyft, to our knolwedge, is the only dataset with 3D box labels along a repeated route, but mostly in nice weather (Palo Alto). Although large-scale, none of these datasets were collected in snowy conditions. Finally, while the Oxford and CADC datasets include data in snowy conditions, they do not include important amodal segmentations, \ie, no labels of invisible and occluded parts of objects. These datasets cannot be used for direct supervised amodal training or evaluation. Both KITTI and \textbf{Cityscapes}~\cite{Cordts2016Cityscapes} have been augmented to have amodal segmentation labels for some classes. \textbf{KINS} \cite{Qi_2019_CVPR} added  labels of traffic participant classes to 14,991 images, and  \cite{lu2020semantic} annotated 500 test images for background classes in Cityscapes and 200 images in KITTI. KINS does not provide the road amodal ground truths addressed in this work, and both KITTI and CityScapes do not have very adverse weather or lighting conditions. There is a critical need in the community for a holistic dataset with amodal labels across adverse weather conditions to be able to train and test the performance and robustness of object detectors, image segmentors, and depth estimators. 

\mypara{Image and amodal segmentation.} 
Image segmentation can be divided into instance segmentation and semantic segmentation. The former aims to identify each individual object instance. Mask R-CNN \cite{he2017mask} is one representative algorithm. The latter aims to label each pixel with a semantic label. State-of-the-art models exploit multi-scale feature fusion (\ie, \cite{zhao2017pyramid}) or self-attention (\eg, \cite{wu2021dannet}).

Although both segmentation tasks have been studied extensively, they only reason about the visible parts of the observed scene and fail to learn the occluded semantic information. Thus, in recent years, amodal segmentation has gained research interest \cite{li2016amodal, Qi_2019_CVPR}. However, due to the lack of amodal labeled datasets, amodal segmentation has mostly focused on learning the invisible information using unsupervised algorithms \cite{li2016amodal}, which  suffer from grave outliers. KINS~\cite{Qi_2019_CVPR} is a pioneer amodal dataset for driving scenes; the paper also proposes an amodal baseline model with an occlusion branch. More recently,  \cite{lu2020semantic} proposes a Semantic Foreground Inpainting (SFI) model from weak supervision using a two-branch (background and foreground) architecture. The model aims to learn the amodal semantic background mask by inpainting the predicted foreground areas with nearby background features. We combine SFI with position and channel attention modules by \cite{wu2021dannet}, to develop a strong baseline for amodal road segmentation. 
\section{Creating the Dataset}
\subsection{The sensor platform}
\label{sec:sensors}
\begin{figure}[h]
\begin{center}
\vspace*{0.5em}
  \begin{subfigure}[t]{\linewidth}
  \centering
  \includegraphics[trim={0 2.9cm 0 2cm},clip,width=0.85\linewidth]{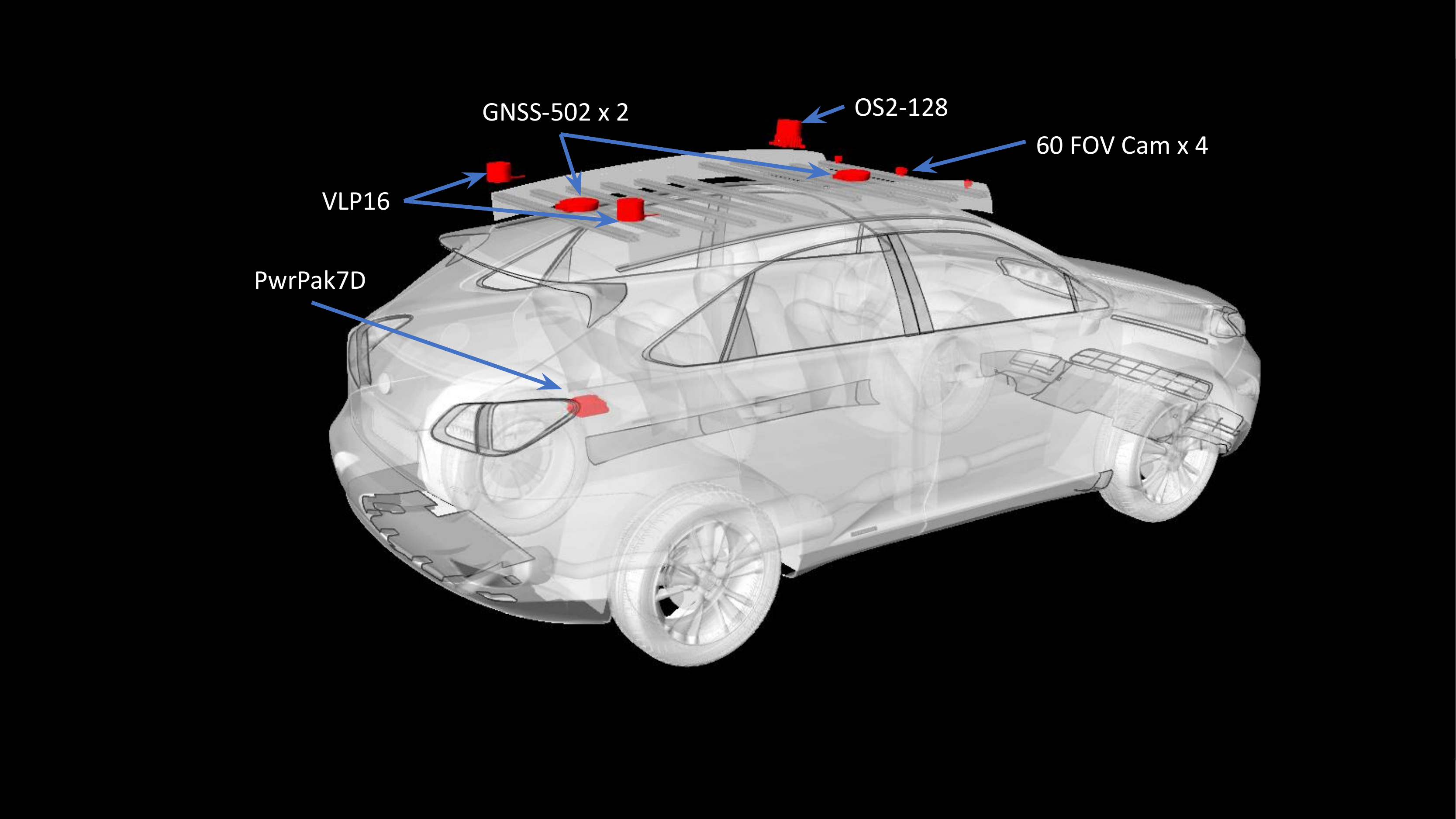}
  \end{subfigure}

  \begin{subfigure}[t]{\linewidth}
  \centering
  \includegraphics[trim={0 2.7cm 0 2cm},clip,width=0.85\linewidth]{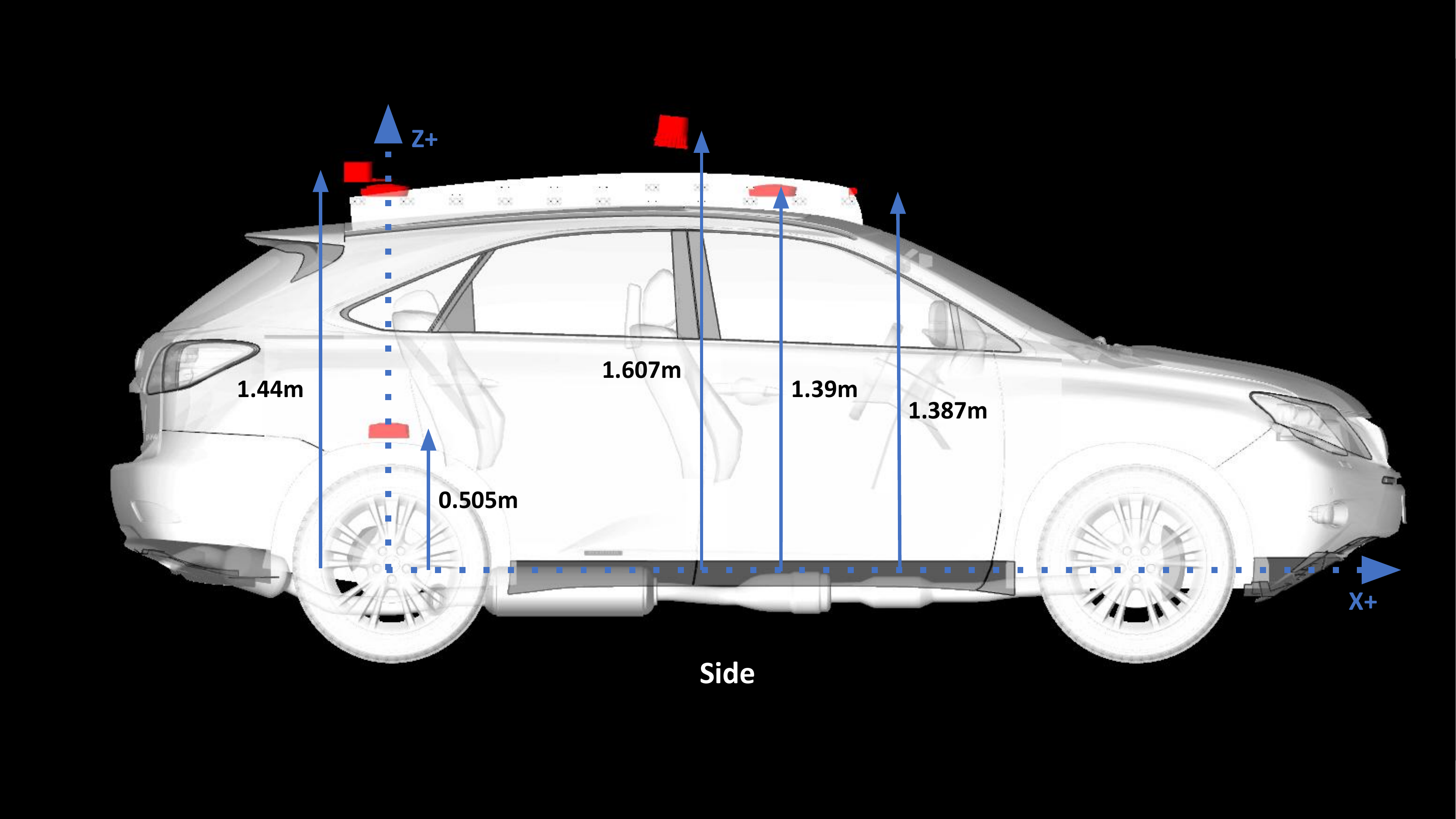}
  \end{subfigure}
  
  \begin{subfigure}[t]{\linewidth}
  \centering
  \includegraphics[trim={0 2.7cm 0 2cm},clip,width=0.85\linewidth]{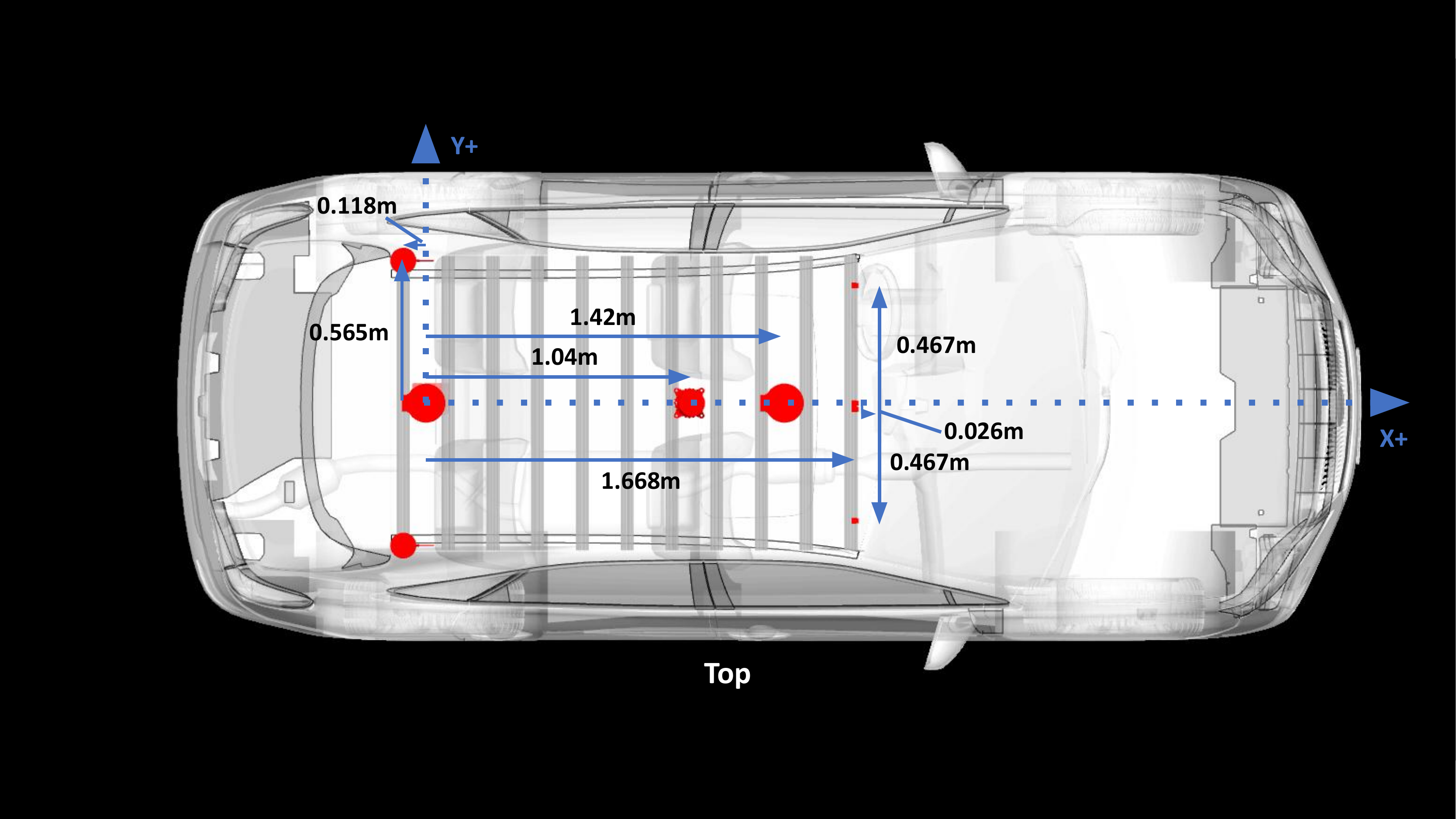}
  \end{subfigure}
  \end{center}
  \vspace{-15pt}
  \caption{\small Sensors and mounting locations on the car.}
  \label{car}
 \vspace{-17pt}
\end{figure}

The data was collected using a self-driving car research platform, illustrated in Figure~\ref{car}. A 2015 Lexus RX 450h is equipped with the following sensors and hardware:
\begin{itemize} [nosep,topsep=3pt,parsep=3pt,partopsep=3pt] 
\item 4$\times$Sekonix SF3325-100 cameras, 1928$\times$1208, 30Hz, GMSL, RCCB Filter, 1/2.7",   60$\degree$ horizontal FOV 
\item 2$\times$Velodyne Puck (VLP-16) LiDAR, 16 scan, 100m range, 30$\degree$ vertical FOV,  10Hz 
\item 1$\times$Ouster OS2-128 long range LiDAR, 128 scan, 240m range, 22.5$\degree$ vertical FOV,  10Hz 
\item 1$\times$Novatel PwrPak7D, dual GNSS-502 antenna, Epson G320N IMU, custom firmware for PTP timing, high precision fixes from PointOne 
\item 1$\times$Nvidia AGX 
\end{itemize}

 The sensors and their locations on the vehicle are shown in Figure \ref{car}. We will release json files that specify extrinsic parameters to each sensor. NVIDIA's recording tool (part of the AGX)   
 was utilized to log the LiDARs at  10 Hz and cameras at 30 Hz. The Novatel GPS/INS data was logged using a PC running ROS, and time synchronized with the AGX through PTP and Novatel custom firmware. Time  synchronization between cameras has been verified to 10s of microseconds. More details on synchronization between sensors have been included in Appendix \ref{ssec:sync_details}. LiDAR scans can be combined with GPS/INS to form accurate 3D reconstructions of the environment as shown in Figure \ref{pointclouds-fig}. 

\subsection{Route and data collection}
\label{sec:route}
\begin{figure}[h]
\centering
\begin{subfigure}[t]{\linewidth}
\centering
    \includegraphics[width=0.9\linewidth]{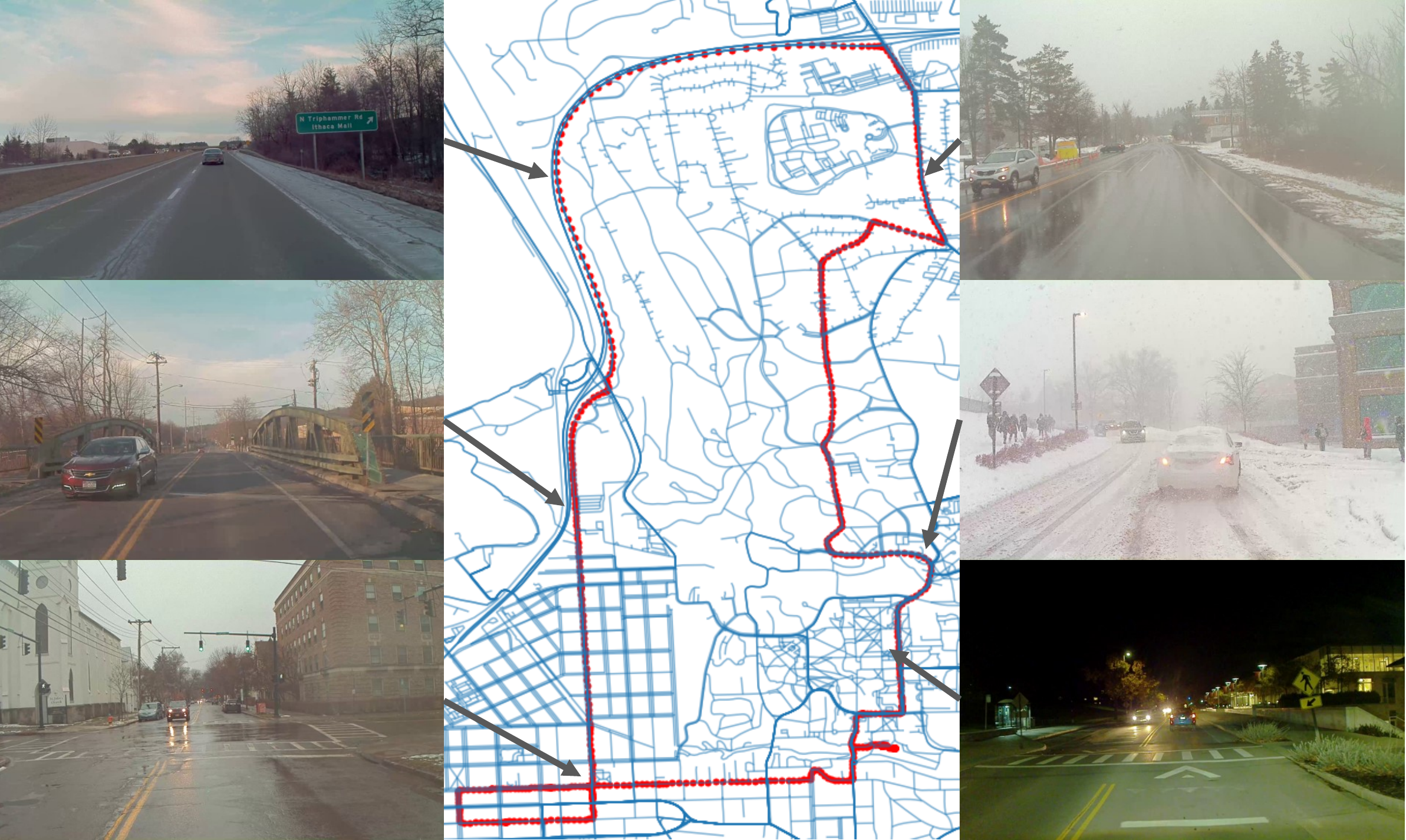}
         \caption{\small Map of the route showing images across different scenes: highway, downtown, rural, suburban, university campus, residential }
         \vskip 5pt
         \label{fig:route and images}
    \end{subfigure}
\begin{subfigure}[t]{\linewidth}
\centering
         \includegraphics[width=0.9\linewidth]{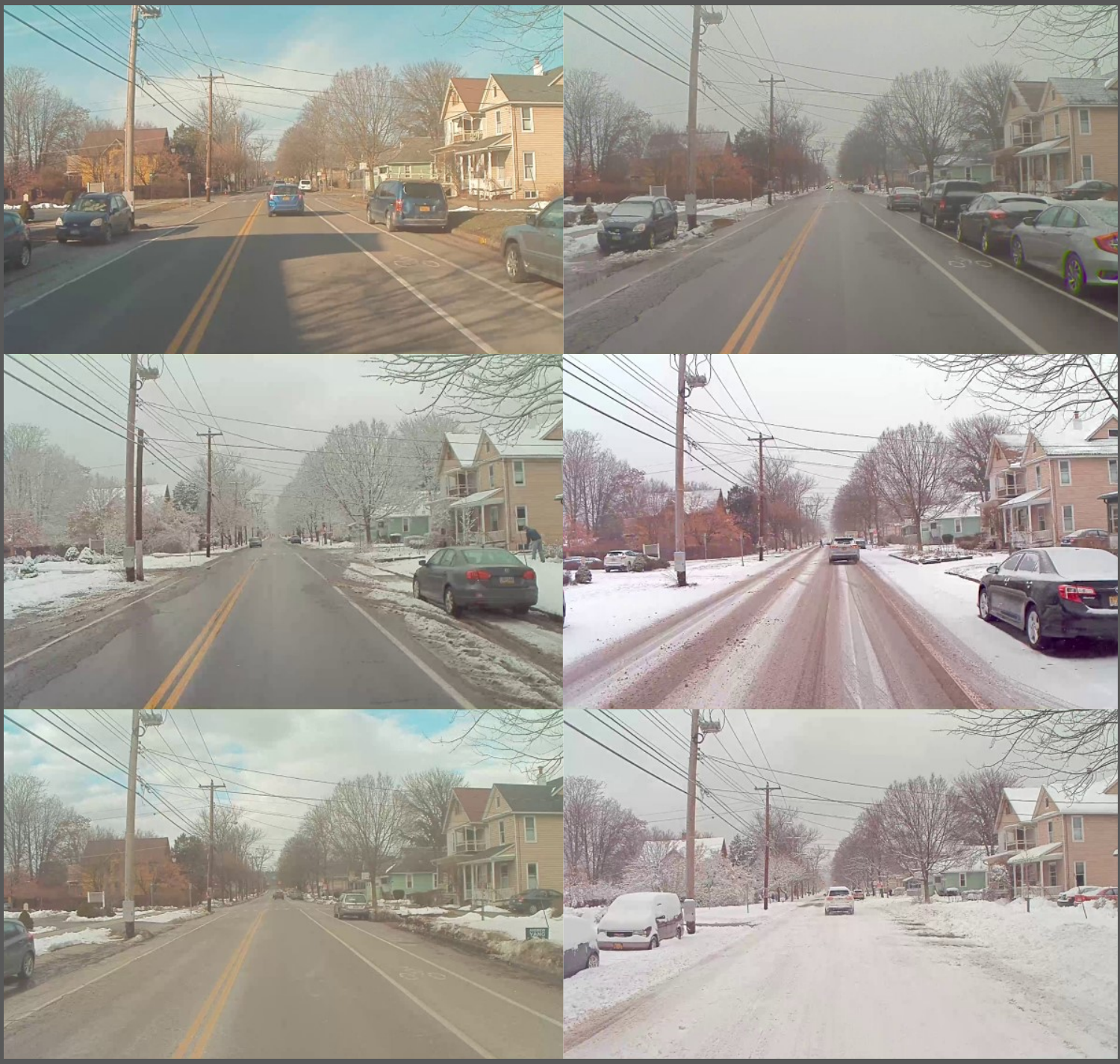}
         \caption{\small Same location across varying weather conditions}
         \label{fig:same location}
\end{subfigure}
\vspace{-10pt}
    \caption{Route and image visualizations}
 \label{overall-route}
 \vskip 5pt
    \centering
 \includegraphics[width=3.25in]{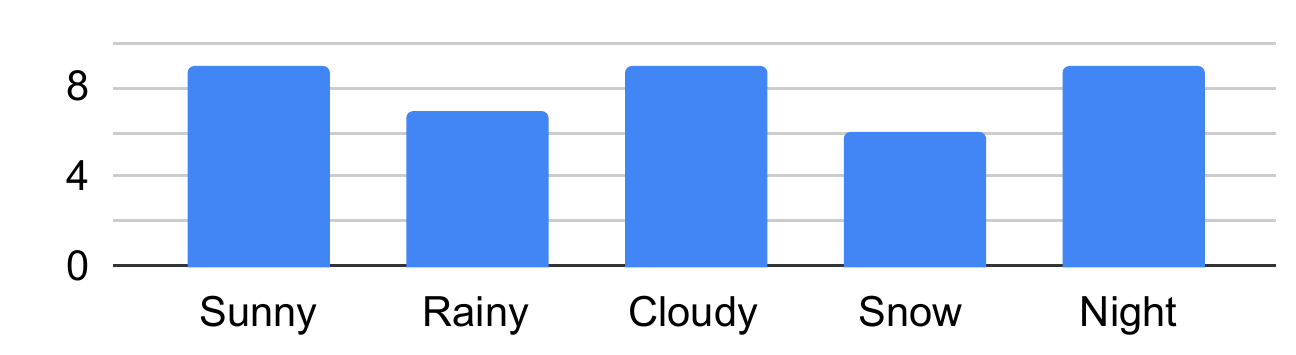}
\vskip -10pt
\caption{ Histogram of conditions present in dataset. }
    \label{fig:weather-folder}
     \vspace{-20pt}
\end{figure}

    A 15 km loop consisting of varying road types and surrounding environments was selected. The route includes university campus, downtown, highway, urban, residential, and rural areas. Figure \ref{overall-route}a shows a map of the route with images captured at several locations. 
    Driving was scheduled to capture data at different times of day, including night. Heavy snow situations were captured before and after roads were plowed. A key uniqueness of our dataset is that the same locations can be observed across different conditions; an example is shown in Figure \ref{overall-route}b.  A breakdown of the traversals under different conditions is shown in Figure \ref{fig:weather-folder}. 
    
\subsection{Sensor calibration}
\mypara{Camera intrinsic, LiDAR to camera extrinsic.} 
For each camera we provide camera intrinsic parameters. The camera intrinsic and distortion parameters were inferred by openCV \cite{opencv_library} camera calibration functions and target checkerboards. The LiDAR to camera extrinsic matrix calibration was performed using multiple checkerboards at different locations/ranges in front of all sensors. Given the camera intrinsic and undistortion parameters, calibration matched the detected target points on checkerboards in LiDAR coordinates with the corresponding target points in camera coordinates. The calibration code along with calibration images will be released for custom analysis. 

\mypara{Lidar to IMU.} We provide calibration between LiDAR and GPS/IMU sensors. LiDAR and GPS/IMU data was collected with large turning angles, and relative transformation (rotation, translation) in the LiDAR  frame was estimated by performing Iterative Closest Point (ICP) between LiDAR point clouds. With the estimated transformation and pose readings from GPS/IMU (in IMU  frame), we use a hand-eye calibration method~\cite{horaud1995hand} to obtain the relative transformation between LiDAR and IMU coordinates.
  Correction for the ego motion within LiDAR sensors has been implemented by interpolating the pose based on the time offset of laser firing during the LiDAR rotation, as specified by  Ouster and Velodyne. This timing can be combined with GPS/INS data to compensate for the ego motion.

\begin{figure*}[h!]
\label{fig:amodal_labels}
  \centering 
  \includegraphics[width=\linewidth]{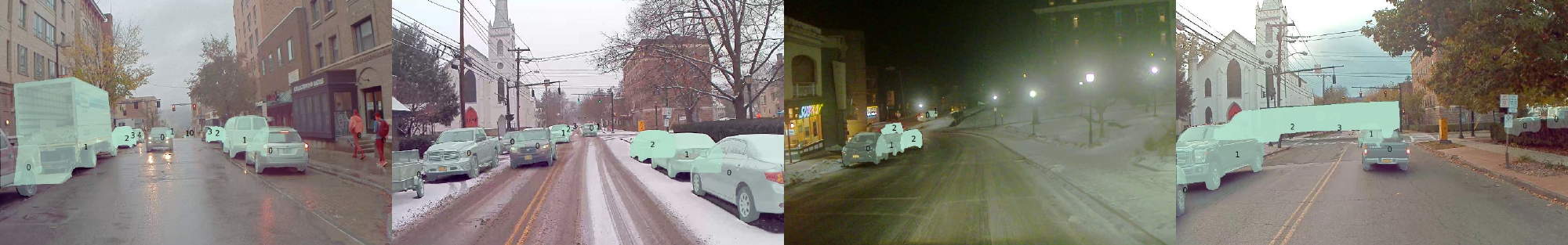}
  \vskip -5pt
  \caption{\small Examples of amodal ground truth masks for cars across different weather conditions.}
  \label{amodal_labels}
  \vspace{-5pt}
\end{figure*}

\begin{figure*}[h!]
  \centering 
  \includegraphics[width=0.7\linewidth]{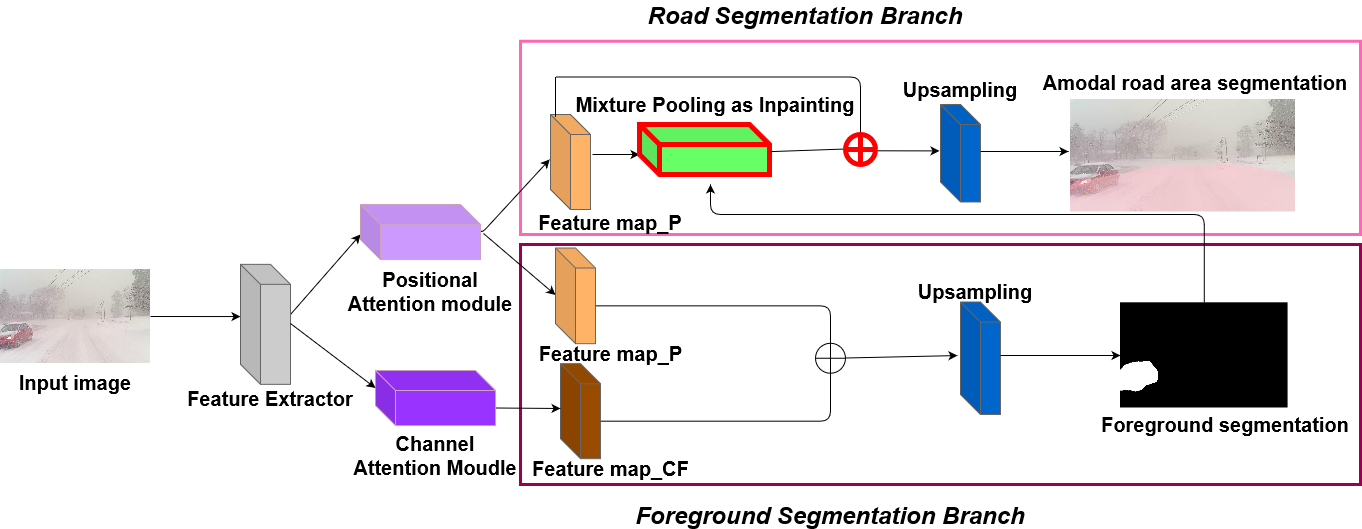}
  \vskip -5pt
  \caption{\small Our proposed amodal road segmentation baseline using PAM, CAM, and modified pooling modules, designed to restore occluded background features. Bold red indicates that the Mixture Pooling as an Inpainting module is added to the road segmentation branch.}
\label{fig:network}
\vspace{-12pt}
\end{figure*}

\subsection{Amodal Labeling: Road}
\label{sec:labeling}
We developed a custom labeling tool to obtain the amodal masks for the road and objects. For road labels across different environmental conditions, such as snow covered roads, \emph{we leveraged the repeated traversals of the same route.} Specifically, \emph{road labels from nice weather are transferred to poor weather conditions, through point cloud road maps built using the GPS pose and LiDAR data.}
The route/data was divided into 76 intervals. The point clouds were then projected into birds-eye-view (BEV) and the road was labeled using a polygon annotator. Once the road has been labeled in the BEV, which yields the 2D road boundary, the height of the road was determined by decomposing the polygon into smaller 150 m$^{2}$ polygons and fitting a plane to the points within the polygon boundary that are within a 1.5 m threshold of the mean height. We then use RANSAC and a regressor to fit a plane to the points; the height for each point along the boundary is then calculated using the estimated ground plane. The road points are projected to the image and a depth mask is created, obtaining the amodal label for the road.  The ground planes can be projected to a particular location of a newly collected route by matching the location to the labeled map with GPS and refining alignment using ICP. A final check for the ICP solution is made by verifying that the mean projected groundtruth mask for the road label meets an 80\% mean IOU with all the other groundtruth masks for the same location, if not the query location data is not retrieved. Ground truth road masks and a labeling example are shown in Appendix \ref{ssec:labeling}.

\subsection{Amodal Instance Segmentations: Objects}
Amodal objects were labeled using  Scale AI for six foreground object classes: car, bus, truck (includes cargo, fire truck, pickup, ambulance), pedestrians, cyclist and motorcyclist. There are three main components in this labeling paradigm: first identifying the visible instances of objects, then inferring the occluded instance segmentation mask, and finally labeling the occlusion order for each object. The labeling is performed on the leftmost front facing camera view. We follow the same standard as KINS \cite{Qi_2019_CVPR}. We have labeled a set of 7000 images across various weathers and locations with split details in Section \ref{exp:amodal_ins}. Example ground truth amodal masks are shown in Figure \ref{amodal_labels}.

\section{Algorithm for Amodal Road Segmentation} 
\label{sec:baselines}
To showcase the environmental diversity and amodal qualities of the dataset, we trained and tested two baseline networks to identify the amodal road at pixel levels, even if the road is covered by snow or cars. 
The first baseline network is \textbf{Semantic Foreground Inpainting (SFI)}~\cite{lu2020semantic}. We show some qualitative results of this model under night, snow, and cluttered environments in Appendix ~\ref{ssec:training_details}. This network fails in conditions with low-light, extreme weather, long-distances, and cluttered pixels. 
We therefore propose a second baseline which improves upon  SFI through the following three innovations in \autoref{fig:network}.

\mypara{Positional and Channel Attention}: Because amodal segmentation is primarily about inferring what is \emph{not} visible, context is an extremely important cue.
Fu et al.~\cite{fu2019dual} introduce two innovations to capture two different kinds of context.
The positional attention module (PAM) uses pixel features to attend to other pixels in the image, in effect capturing context from other parts of the image.
The channel attention module uses a similar attention mechanism, effectively aggregating information over \emph{channels}.
We apply both these modules over the backbone feature extractor as shown in \autoref{fig:network}. (Details of these modules are in Appendix \ref{ssec:baselines}). We combine CAM and PAM to better locate the fine mask boundary. The final foreground instance mask is obtained by an up-sampling layer.

\mypara{Mixture Pooling as Inpainting}: The Max Pooling as Inpainting operation proposed by~\cite{lu2020semantic} can be used to help restore the amodal road features by replacing overlapping foreground features with nearby background features. However, since the background features are usually smoothly distributed, a max pooling operation will be very sensitive to any noise added to them. In contrast, an average pooling operation could naturally mitigate the noise. To this end, we propose to combine average pooling and max pooling for inpainting, which we name Mixture Pooling. We show some qualitative results in Appendix \ref{ssec:baselines}: the Mixture Pooling module produces foreground masks of smoother boundaries than the original max pooling operation.

\mypara{Sum operation}: Before the final up-sampling layer, instead of directly passing the features from the mixture pooling module, we include a residual link from the output of the PAM module.
By optimizing the two feature maps jointly in the road segmentation branch, the PAM module can also learn the background features in occluded regions. This can lead to a more accurate restoration of the background features.  We include qualitative results in Appendix ~\ref{ssec:baselines} to demonstrate that the sum operation produces better inpainting results than the network's performance without the sum operation, especially when the foreground objects (\textit{e.g.}, cars) occupy a relatively large space in the images.

\section{Experimental Baselines and Results}

\subsection{Amodal Road Segmentation}
\label{exp:amodal_road}
\mypara{Dataset splits for road.} One of the main goals of this dataset is to assess the performance of amodal road segmentation across weather conditions. To create a  training, validation, and test set, we pick 590 locations throughout the route and retrieved images and road depth masks for each of these locations for 25 different traversals. After filtering out locations with poor alignment after ICP (as explained in the labeling section), we have a dataset with 11,475 images for training and  3,275 for testing. \emph{We ensure that the locations in testing are not visible from the locations in training.} Additionally, we further split this dataset by weather condition, creating 5 smaller datasets across all weather conditions (sunny, rainy, cloudy, snowy, and night). For each condition, we include the same number of images for each location. These smaller datasets are meant to study the effect of domain gaps between weather conditions, given that all datasets have the same amount of images/data and have seen the same locations.  Details of the number of images in our dataset along with other published test sets from KITTI road~\cite{Fritsch2013ITSC} and Cityscapes~\cite{Cordts2016Cityscapes} are shown in Table~\ref{table:config}.
\begin{table}

\captionof{table}{Dataset splits (\# of images) of amodal road segmentation.}
\vspace*{-0.5em}
\small
\centering
\begin{tabular}{lp{1.15cm}p{1.15cm}}
\toprule
Dataset            & Train Set & Test Set \\ \midrule
Cityscapes~\cite{Cordts2016Cityscapes}          & 2,975               & 500               \\ 
KITTI~\cite{Fritsch2013ITSC}             & N/A                & 200               \\ 
Ithaca365 (All)         & 11,475              & 3,275              \\ 
Ithaca365 (each weather) & 2,295                & 655               \\ 
\bottomrule
\end{tabular}
\label{table:config}
\vspace{-11pt}
\end{table}

\begin{figure}[t]

  \centering 
  \includegraphics[width=\linewidth]{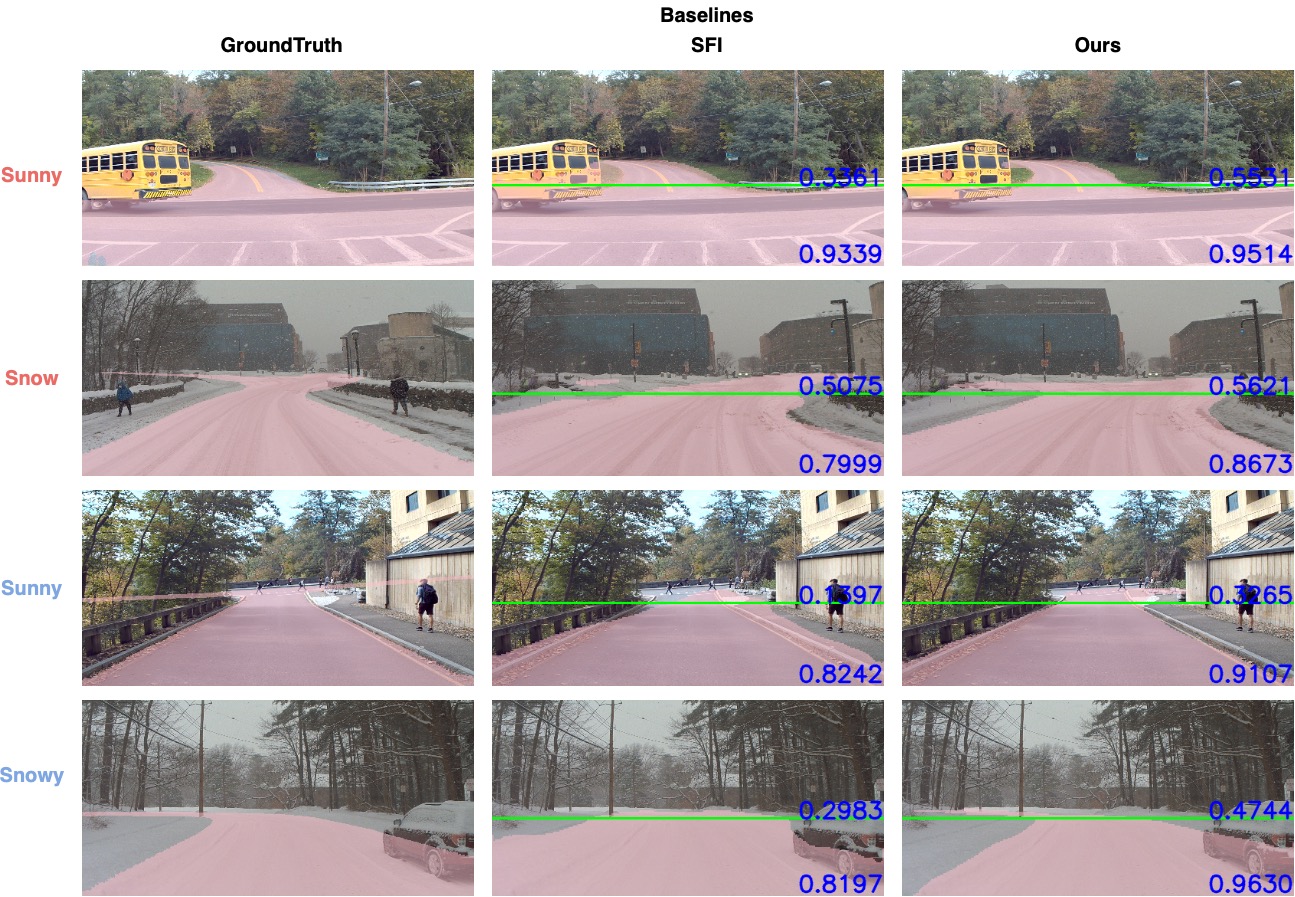}
  \caption{\small Amodal road segmentation results for the two baselines (SFI, OURS). Top two rows are trained on sunny conditions, with the first row tested on sunny and second on snowy. Bottom two rows are trained on snowy conditions, with the third row tested on sunny and the fourth on snowy. The cutoff depth of 30m is marked with a green line, with the \textit{far} IOU above and \textit{close} IOU below.}
  \label{fig:near_and_far}
  \vspace{-17pt}
\end{figure}

\mypara{Evaluation metrics for road.}
Segmentation tasks are typically evaluated through mean IOU. However, we find that this metric is not appropriate for amodal road segmentation. The vast majority of the road pixels in an image correspond to the first few meters in front of the car, which are easy to segment. We find that our baselines often achieve a high IOU but only correctly segment the immediate vicinity of the ego-car (Figure \ref{fig:near_and_far}). However, for self-driving, accurate segmentation of far-away pixels is needed, as they influence planning and future decisions. 

To ensure that algorithms do not ignore far-away road pixels, we introduce a metric that divides the evaluation into two bins, \textit{close} and \textit{far}, by using the depth mask. We denote a range of 30m or less as \textit{close} and anything above as \textit{far}. This follows a similar custom to other datasets such as KITTI, that bin object detection, as \textit{easy}, \textit{medium}, or \textit{hard} based on distance. In Table~\ref{tbl:5x5} and Figure~\ref{fig:near_and_far}, we quantitatively and qualitatively demonstrate the need for an evaluation that is based on distance. Quantitatively, we can see that the model performance is high and does not vary much when looking at close ranges and even across weather conditions. Qualitatively, in Figure \ref{fig:near_and_far}, we can see that below the threshold distance the model captures the road well, but above it, branches or far away roads are not captured. 

\begin{table}[t]
\setlength{\tabcolsep}{1.8pt}
\footnotesize
\renewcommand{\arraystretch}{1.00}
\centering
\caption{\small \textbf{Amodal road results (IOU for road) among five weather conditions.} We report training on \emph{row} and testing on \emph{column}, using the SFI/OURS models. Diagonal entries are the in-domain model. The best result for each column is \textbf{bold}. \label{tbl:5x5}}
\vspace*{-1em}
\begin{tabular}[b]{c|p{1.2cm}p{1.2cm}p{1.2cm}p{1.2cm}p{1.2cm}}

    \toprule
    
    Far & \multicolumn{1}{c}{Sunny} & \multicolumn{1}{c}{Cloudy} & \multicolumn{1}{c}{Rainy} & \multicolumn{1}{c}{Snowy} & \multicolumn{1}{c}{Night} \\ \midrule
    Sunny & \textbf{49.0}/\hspace{0.02em}{ \textbf{57.3}} & 44.0/\hspace{0.02em}{ {51.7}} & 44.6/\hspace{0.02em}{ {49.8}} & 40.6/\hspace{0.02em}{ {47.0}} & 22.0/\hspace{0.02em}{ {46.7}} \\
    Cloudy   &37.8/\hspace{0.02em}{ {48.8}}& {\textbf{45.8}/\hspace{0.02em}{ \textbf{{58.0}}}} & 37.1\hspace{0.02em}/\hspace{0.02em}{ {49.1}} & 36.2\hspace{0.02em}/\hspace{0.02em}{ {44.1}} & 26.0\hspace{0.02em}/\hspace{0.02em}{ {31.2}} \\
    Rainy   & 39.0\hspace{0.02em}/\hspace{0.02em}{ {46.6}}& 45.1\hspace{0.02em}/\hspace{0.02em}{ {53.8}}& \textbf{46.5}/\hspace{0.02em}{ \textbf{{55.5}}} & 37.9\hspace{0.02em}/\hspace{0.02em}{ {46.3}} & 29.7/\hspace{0.02em}{ {45.0}} \\
    Snowy   & 40.5\hspace{0.02em}/\hspace{0.02em}{ {44.9}}& 41.4\hspace{0.02em}/\hspace{0.02em}{ {52.0}} & 38.8\hspace{0.02em}/\hspace{0.02em}{ {49.6}}& {\textbf{44.0}/\hspace{0.02em}{ \textbf{{54.6}}}} & 35.4\hspace{0.02em}/\hspace{0.02em}{ {46.6}} \\
    Night  & 31.1\hspace{0.02em}/\hspace{0.02em}{ {43.7}} & 36.6\hspace{0.02em}/\hspace{0.02em}{ {50.3}} & 33.9\hspace{0.02em}/\hspace{0.02em}{ {48.5}} & 30.4\hspace{0.02em}/\hspace{0.02em}{ {52.4}}& {\textbf{38.2}}/\hspace{0.02em}{ \textbf{{55.1}}} \\\midrule
   Close & \multicolumn{1}{c}{Sunny} & \multicolumn{1}{c}{Cloudy} & \multicolumn{1}{c}{Rainy} & \multicolumn{1}{c}{Snowy} & \multicolumn{1}{c}{Night} \\ \midrule
    Sunny & \textbf{91.6}/\hspace{0.02em}{\textbf{{95.5}}} & 89.4/\hspace{0.02em}{ {92.0}} & 86.5/\hspace{0.02em}{ {88.4}} & 85.7/\hspace{0.02em}{ {88.7}} & 82.8/\hspace{0.02em}{ {90.0}} \\
    Cloudy   &91.2/\hspace{0.02em}{ {85.6}}& {\textbf{91.9}/\hspace{0.02em}{ \textbf{{95.7}}}} & 83.0\hspace{0.02em}/\hspace{0.02em}{ {86.8}} & 83.4\hspace{0.02em}/\hspace{0.02em}{ {88.5}} & 82.9\hspace{0.02em}/\hspace{0.02em}{ {86.8}} \\
    Rainy   & 89.1\hspace{0.02em}/\hspace{0.02em}{ {90.7}} & 87.7\hspace{0.02em}/\hspace{0.02em}{ {89.0}}& \textbf{89.7}/\hspace{0.02em}{ \textbf{{94.9}}} & 85.6\hspace{0.02em}/\hspace{0.02em}{ {88.3}} & 85.3/\hspace{0.02em}{ {89.2}} \\
    Snowy   & 86.8\hspace{0.02em}/\hspace{0.02em}{ {89.0}} & 90.7\hspace{0.02em}/\hspace{0.02em}{ {92.2}} & 84.3\hspace{0.02em}/\hspace{0.02em}{ {86.5}}& {\textbf{91.2}/\hspace{0.02em}{ {\textbf{92.9}}}} & 89.3/\hspace{0.02em}{ {91.5}} \\
    Night  & 87.5\hspace{0.02em}/\hspace{0.02em}{ {90.3}} & 90.5\hspace{0.02em}/\hspace{0.02em}{ {93.6}} & 85.4\hspace{0.02em}/\hspace{0.02em}{ {87.2}} & 89.4\hspace{0.02em}/\hspace{0.02em} {92.0}& \textbf{91.3}/\hspace{0.02em}{ {\textbf{93.6}}} \\\bottomrule
\end{tabular}
\label{tbl:aroad}
\vspace{-12pt}
\end{table}

\mypara{Experimental results.}
We perform experiments on our dataset, along with the publicly released training/test sets for CityScapes and KITTI  \cite{lu2020semantic}. Before the two baseline models (see Section~\ref{sec:baselines}) are trained using our amodal road dataset, we generate the required foreground semantic masks (\textit{i.e.} cars, trucks, buses, and persons) for both baselines using a
pretrained DeepLab~\cite{chen2017deeplab} network. Our baseline model is trained using two GPUs (NVIDIA GT 1080 Ti) in parallel and the SFI is trained using one. Both of the networks are trained using Adam optimizer with a batch size of $8$. The overall experimental results are shown in Table~\ref{tab:total_results}. 

\begin{table}[t]
\small\centering
\caption{ Amodal road segmentation results (mIOU) for SFI/OURS models trained/tested on Cityscapes, KITTI, and Ithaca365.}
\resizebox{.45\textwidth}{!}{%
\begin{tabular}{lc|ccc}
\toprule
\multicolumn{1}{c}{\multirow{2}{*}{Model}} & \multicolumn{1}{c|}{\multirow{2}{*}{Train Set}} & \multicolumn{3}{c}{IOU on Test Set }                     \\   \cmidrule{3-5}
&                           & KITTI~\cite{Fritsch2013ITSC} &   Cityscapes~\cite{Cordts2016Cityscapes} & Ours  

                               \\ \midrule
SFI\cite{lu2020semantic}                  & Cityscapes                &    66.29        &  72.75   & 50.16                                      \\ 
SFI\cite{lu2020semantic}                     & Ithaca365      &  \textbf{72.31}     &      \textbf{74.68}  & \textbf{89.50}                                     \\\midrule
Ours                     & Ithaca365      &  72.25     &      77.43  &  92.19                                    \\ \bottomrule
\end{tabular}%
}
\vspace{-13pt}
\label{tab:total_results}
\end{table}

We observe that models trained on our dataset are in general better than those trained on Cityscapes. For instance, when tested on ours, the SFI model trained on ours significantly outperforms that trained on Cityspaces (89.50 vs. 50.16). While this may not be surprising, as in-domain results are normally better than out-domain results, reversely when tested on Cityspaces, the SFI model trained on ours is still better than that trained on Cityspaces (74.68 vs. 72.75). Moreover, when tested on KITTI, the SFI model trained on ours notably outperforms that trained on Cityspaces. These results together suggest that our dataset is more useful for training a robust and generalizable model. Also, we want to note that the 2,975 Cityscapes images used in SFI [24] have normal modal segmentation masks thus they are weak labels. Additionally for the test set, the authors of SFI manually labeled 500 images with the amodal masks for three background classes (\textit{sidewalk, road, other rigid world}). Due to the difficulty of annotating amodal masks based on one image especially for far away and occluded pixels, the annotations are coarse and are prone to errors. We believe weak labels and noisy ground truths lead to the large gap in model performance between the Cityscapes and Ithaca365. Across models (SFI vs. ours), we find that our proposed model is better than prior art.

We also evaluate the SFI and our baseline models across 5 different weather conditions (Table~\ref{tbl:aroad}) by using our proposed Far and Close IOU metrics . In close quarters, both methods perform well, yet large performance gaps are found for pixels farther than the 30m threshold --- either between SFI and our models, or between in-domain and cross-domain results. This indicates that 1) there are significant domain gaps between different weather/time conditions, and revealing these gaps requires stratifying evaluations by distance; 2) our model is more robust than the SFI model. Snowy and night appear to be the most challenging weather conditions for this task since the models that are not trained under these conditions perform worse under these two conditions. One potential reason for this phenomena is that snowy and night have the most visual degradation among all weather conditions, which cause obstacles during the feature extraction stage for both models. 
\label{sec:exp}

\subsection{Amodal Instance Segmentation}
\label{exp:amodal_ins}
Our second set of experiments is on amodal instance segmentation, using our annotated 7,000 images of six classes.
We adopt Mask R-CNN~\cite{he2017mask,Detectron2018} as the baseline model. We conduct two experiments. First, we train the model using our overall amodal training set, which consists of 5,600 images collected across five different weather conditions. We then evaluate the model on our overall amodal test set, which consists of 1,400 images (Table~\ref{tbl2:instance}). The training and test data are strictly from different locations. We achieve a decent overall mean average precision (mAP) of 56.5. 

Second, we split the dataset into five different weather conditions (sunny, snowy, cloudy, rainy and night). On each of the five weather conditions we prepare train and test sets with different quantities of images as different weathers vary in frequency across the 40 labeled traversals. In general, the train set has 1120 images and the test set has 280 images but rainy and snow conditions have less. Then for each weather condition, we train Mask R-CNN models respectively and evaluate the mAP of the trained model on the test set of all five weather conditions. As shown in Table~\ref{tbl2:instance}, for all five weather conditions, training and testing on the same weather condition has the highest mAP. This confirms that that domain adaption is a challenging task for Amodal instance Segmentation. We still observe that snowy, night and rainy have lower mAP than sunny and cloudy, meaning that they are more challenging cases and deserve further, more focused research efforts. Visualizations of the amodal instance segmentation are included in Appendix ~\ref{ssec:training_details}.

\begin{table}[t]

\small
\centering
\caption{\small \textbf{Amodal instance segmentation results (mAP@[.5:.95]) across five weather conditions, using Mask R-CNN~\cite{he2017mask}}. Each entry  means training on \emph{row} and testing on \emph{column}. The best result for each column is \textbf{bold}.\label{tbl2:instance}}
\vspace*{-0.5 em}

\begin{tabular}{c|c|c|c|c|c}
\toprule
Train \textbackslash Test  & Sunny & Cloudy & Rainy & Snowy & Night \\ \hline
Sunny & \textbf{54.3} & 47.6 & 38.9 & 41.1 & 29.4 \\ \hline
Cloudy & 47.2 &\textbf{ 48.0} & 33.6 & 21.4 & 14.2  \\ \hline
Rainy & 35.5 & 32.1 & \textbf{46.3} & 33.2 & 21.0 \\ \hline
Snowy & 33.9 & 40.4 & 35.8 & \textbf{47.0} & 21.4 \\ \hline
Night & 26.7 & 27.7 & 24.9 & 23.7 & \textbf{30.2} \\ \bottomrule
\end{tabular}
\vspace{-7pt}
\end{table}

\begin{table}[!t]
	\tabcolsep 2pt
	\renewcommand{\arraystretch}{1.05}
	\centering
	\caption{\textbf{3D car detection result on our dataset, using pre-trained Point R-CNN~\cite{shi2019pointrcnn} on \kitti and \argo}. We  report  \APBEV/\AP of  the car category  at  IoU= 0.5/0.7 across different  depth  ranges  on  the test set.
	\label{tbl:kitticc2kittirr}}
	\vspace{-.5em}

\resizebox{.47\textwidth}{!}{%
\begin{tabular}{=l|+C{43pt}|+C{43pt}|+C{43pt}|+C{43pt}|+C{43pt}|+C{43pt}}
    \hline
     & \multicolumn{3}{c|}{IoU 0.5} & \multicolumn{3}{c}{IoU 0.7} \\ \cline{2-7}
    \multicolumn{1}{c|}{Model}            & 0-30    & 30-50    & 50-80 & 0-30    & 30-50    & 50-80   \\ \hline
     KITTI & 64.2 / 61.8 & 45.3 / 42.7 & 26.3 / 20.8 &   41.5 / 26.0 & 32.3 / 13.1 & 10.2 / \phantom{0}5.3   \\ 
    Argo & 54.2 / 53.1 & 33.6 / 27.5 & 15.4 / 13.8  & 40.2 / 15.9 & 18.3 / \phantom{0}6.6 & 13.3 / \phantom{0}9.1 \\
    \hline

    \end{tabular}%
}
    \vspace{-10pt}
\end{table}

\begin{figure}[!t]
    \centering
    \includegraphics[width=.9\linewidth]{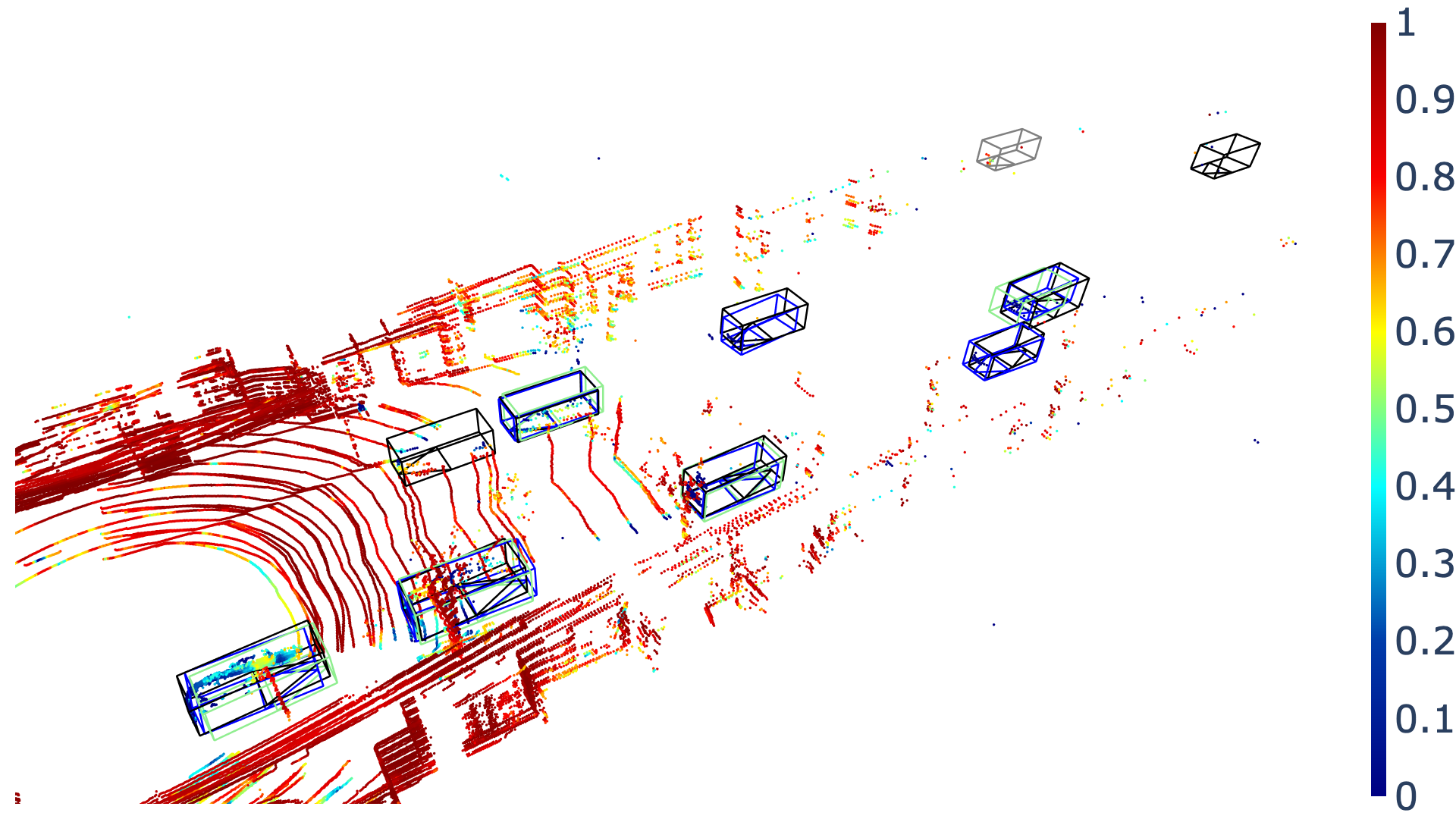}
    \vspace{-10pt}
    \caption{\textbf{Ephemerality from multiple traversals.} With observations from multiple traversals, we compute the \emph{ephemerality} statistics introduced in \cite{barnes2018driven}, and color the point cloud by the scale. We show ground-truth labels in black, true positive detections from models trained on \kitti in green, on \argo in blue, and false positive ones in gray.
    Multiple traversals enable easy false positive filtering by thresholding the ephemerality.
    }
    \label{fig:pp_score_vis}
    \vspace{-15pt}
\end{figure}
\subsection{3D Object Detection, Repeated Routes}
Our third set of experiments is on 3D object detection using LiDAR data. For this experiment a small test set consisting of 175 point clouds at different locations, and annotated 3D bounding boxes for six classes (car, truck, bus, cyclist, pedestrian) is used. We then evaluate a LiDAR detector, Point-RCNN \cite{shi2019pointrcnn}, pre-trained on \kitti and \argo.  Point-RCNN is a popular method which incorporates PointNet \cite{qi2017pointnet} point wise features and faster R-CNN \cite{ren2015faster} to directly generate proposals in 3D using only LiDAR points. 

We follow \kitti to report average precision (AP) in 3D and bird-eye-view (BEV), denoted by \AP and \APBEV, respectively. We split AP by depth ranges. We focus on the \emph{Car} category as it is the most commonly observed class. We report \emph{Car} AP with the intersection over union (IoU) thresholds at 0.5 or 0.7, meaning, a ground truth car is correctly detected if the IoU between it and the detected box is larger than 0.5 or 0.7.  The result is shown in Table \ref{tbl:kitticc2kittirr}. This serves as a baseline to further study the use of our dataset for 3D object detection across weather conditions and repeated routes. We further demonstrate the \emph{ephemerality} statistics introduced in \cite{barnes2018driven}, which make use of multiple traversals of the same location to determine if a point is ephemeral, \ie, not persistent (see \Cref{fig:pp_score_vis}). This information can be used to remove false positives to improve a detector's mAP, if we have traversed the same location multiple times earlier. 
\vspace{-7pt}
\subsection{Stereo Disparity Estimation}
Our final experiment is on stereo disparity estimation to further showcase the challenge of diverse weather and lighting conditions of our dataset. For each of the conditions (sunny, snow, rain, and night), we collect 4,739/1,188 left-right image pairs (the baseline is 0.46 m) for training and testing. The LiDAR data used as depth ground truth in this experiment is from a previous sensor setup where  4 VLP-16s were mounted in front of the vehicle but the cameras remain unchanged. We obtain sparse ground-truth depths (and so pixel disparities) for pixels that have corresponding LiDAR points. We then train and test a baseline model of stereo matching PSMNet~\cite{chang2018pyramid} on each condition. 

\begin{figure}[t]
  \centering 
  \includegraphics[width=\linewidth]{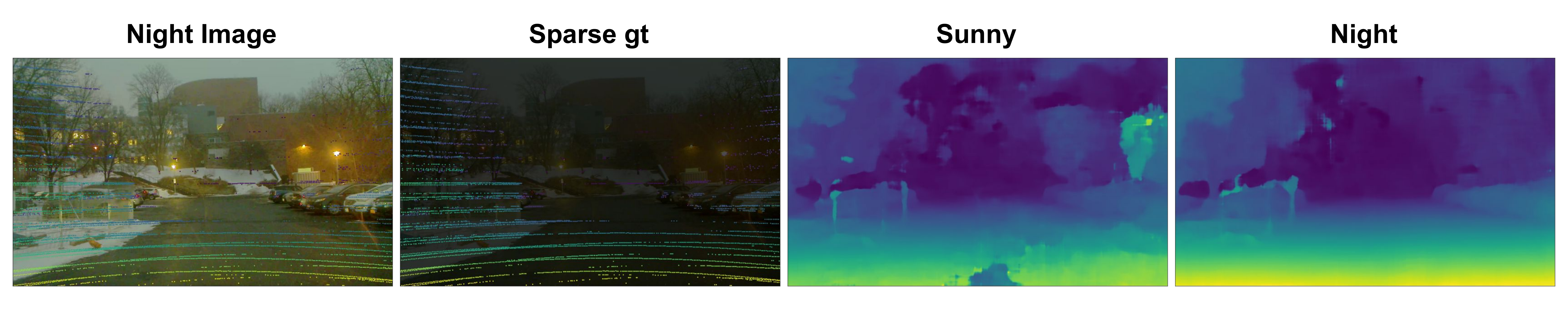}
  \caption{\small Qualitative results of disparity maps, using PSMNet~\cite{chang2018pyramid} trained on different weather conditions.}
  \label{fig:stereo}
  \vspace{-13pt}
\end{figure}

Table~\ref{tbl3:psm} shows the results. We again observe a large domain gap, even in the low-level vision task of stereo matching: the diagonal in-domain errors are usually much smaller than other cross-domain errors of the same columns. This can also be seen qualitatively in Figure~\ref{fig:stereo}: on a night-time image, PSMNet trained using night images notably outperforms using sunny images. Besides, by comparing different columns (different testing cases), we clearly see the difficulty in performing stereo disparity matching at rainy, snowy, and night conditions than the sunny condition. All these results again suggest the timely need of a dataset like ours, with more diverse weather conditions.

\label{sec:exp}
\label{sec:depth-est}
\begin{table}[t]
\small
\centering
\caption{\small\textbf{Three-pixel-error (\%) of stereo matching task using PSMNet \cite{chang2018pyramid} among four different conditions.} On each entry (\emph{row}, \emph{column}), we report training on \emph{row} and testing on \emph{column}.  \label{tbl3:psm}}
\vspace*{-0.5 em}
\begin{tabular}{l|c|c|c|c}
\toprule
Train \textbackslash Test & Sunny & Rain & Snow & Night \\ \hline
Sunny                     & 9.48                     & 23.58                   & 24.41                    & 17.79     \\ \hline
Rain                      & 12.42                      & 24.05                    & 27.45                    & 19.90    \\ \hline
Snow                      & 13.01                    & 27.84                    & 20.99                    & 22.83     \\ \hline
Night                     & 10.58                     & 23.6                    & 24.46                    & 11.99    \\ \bottomrule
\end{tabular}%
\vspace{-15pt}
\end{table}

\begin{table}[t]
\caption{\small Table of train/test data split we used for amodal road segmentation, instance segmentation and 3D object detection tasks. The first column indicates the total frames in our dataset.
}
\vspace*{-0.5 em}
\small
\begin{tabular}{@{}l|l|l|l}

\hline
                                        & Total & Train & Test  \\ \hline
\;\#frames                                & 682217  &   N/A  & N/A    \\ \hline
\;\#frames w/ amodal road masks     &   14750  & 11475 & 3275 \\ \hline
\;\#frames w/ amodal instance masks &   7000  & 5600   & 1400   \\ \hline
\;\#pcls w/ 3D bounding boxes             &   175   &   N/A   & 175    \\ \hline
\end{tabular}
\label{tbl:number}
\vspace{-17pt}
\end{table}
\section{Conclusion, Discussion, Data Release Plan}
\vspace{-5pt}
\label{sec:Discussion}
Current challenges in self-driving car perception are lagging due to the lack of datasets with poor weather, varying light, and dense traffic conditions. This paper develops a unique dataset for the research community, with amodal road and object mask labels across varying scenes (university campus, downtown, highway, urban, residential, and rural) and environmental conditions (snow, rain, night). By using data across repeated traversals of the same route, we can efficiently create a large scale, diverse dataset with amodal labels. The repeated routes also opens potential new research directions such as object discovery and continual learning. A new metric is introduced for the proper evaluation of amodal segmentation of background classes such as road. Baseline road detection, depth estimation, instance segmentation, and 3D detection using this dataset reveals significant performance gaps under varying weather conditions, demonstrating the usefulness of this dataset.

A key advantage of our dataset is the availability of multiple traversals over the same route. These repeated traversals capture the reality that many people repeatedly drive the same routes between home, work and leisure. Information from past traversals can not only improve detections\cite{anonymous2022hindsight} but also allow one to distinguish between static background and mobile objects in the foreground without supervision\cite{barnes2018driven}.
As such, our dataset is a useful testbed for \emph{weakly supervised/unsupervised 3D object discovery and detection}. 
We argue that research on this problem is the need of the hour: current reliance on costly human-annotated data is limiting.

\mypara{Data release plan:} 
Ithaca365 contains \emph{aligned} LiDAR point clouds as well as images captured across 40 traversals of our 15km route. We will provide ground-truth 3D object bounding boxes and amodal masks for six object classes as well as ground truth amodal masks on 7K selected frames. These numbers are similar to the KITTI benchmark~\cite{geiger2012we}. The validation data is carefully selected from different geo-locations from the training data, on different days. We will continue growing the dataset even after this initial release by collecting data over future seasons. IRB deemed this project not Human Participant Research therefore does not require IRB approval.
\vspace{-15pt}
\section{Acknowledgements}
\vspace{-8pt}
\label{sec:Acknowledgements}
This research is supported by grants from NSF (IIS-1724282, IIS-2107161, IIS-2107077), ONR (N00014-17-1-2175), and SRC: 2019-AU-2891. We also thank Scale AI, Amazon (\href{https://aws.amazon.com/sagemaker/data-labeling/}{Ground Truth Plus}), and Point One Navigation. 

{
    \small
    \bibliographystyle{ieee_fullname}
    \bibliography{macros,main}
}

\appendix

\setcounter{page}{1}

\twocolumn[
\centering
\Large
\vspace{0.5em}Supplementary Material \\
\vspace{1.0em}
] 
\section*{\LARGE Supplementary Material}
\pagestyle{empty}
\section*{Appendix}
We provide details omitted in the main text.

\begin{itemize}
    \item \autoref{sec:data_details}: Dataset details. Sensor time synchronization (cf. \autoref{sec:sensors} of the main paper). Road labeling pipeline and additional visualizations of generated depth mask ground-truth (cf. \autoref{sec:labeling} of the main paper). Dataset class statistics.
	\item \autoref{ssec:baselines}: Additional details on baseline algorithms (cf. \autoref{sec:baselines} of the main paper).
	\item \autoref{ssec:training_code}: Link to source code for training and inference (cf. \autoref{exp:amodal_road} of the main paper).
	\item \autoref{ssec:training_details}: Training details and visualization results (cf. \autoref{exp:amodal_road} and of \autoref{exp:amodal_ins} the main paper).
\end{itemize}

\section{Dataset details}
\label{sec:data_details}
\subsection{Visualizations of depth mask ground-truth}
\begin{figure}[ht]

  \centering 
  \includegraphics[width=1\linewidth]{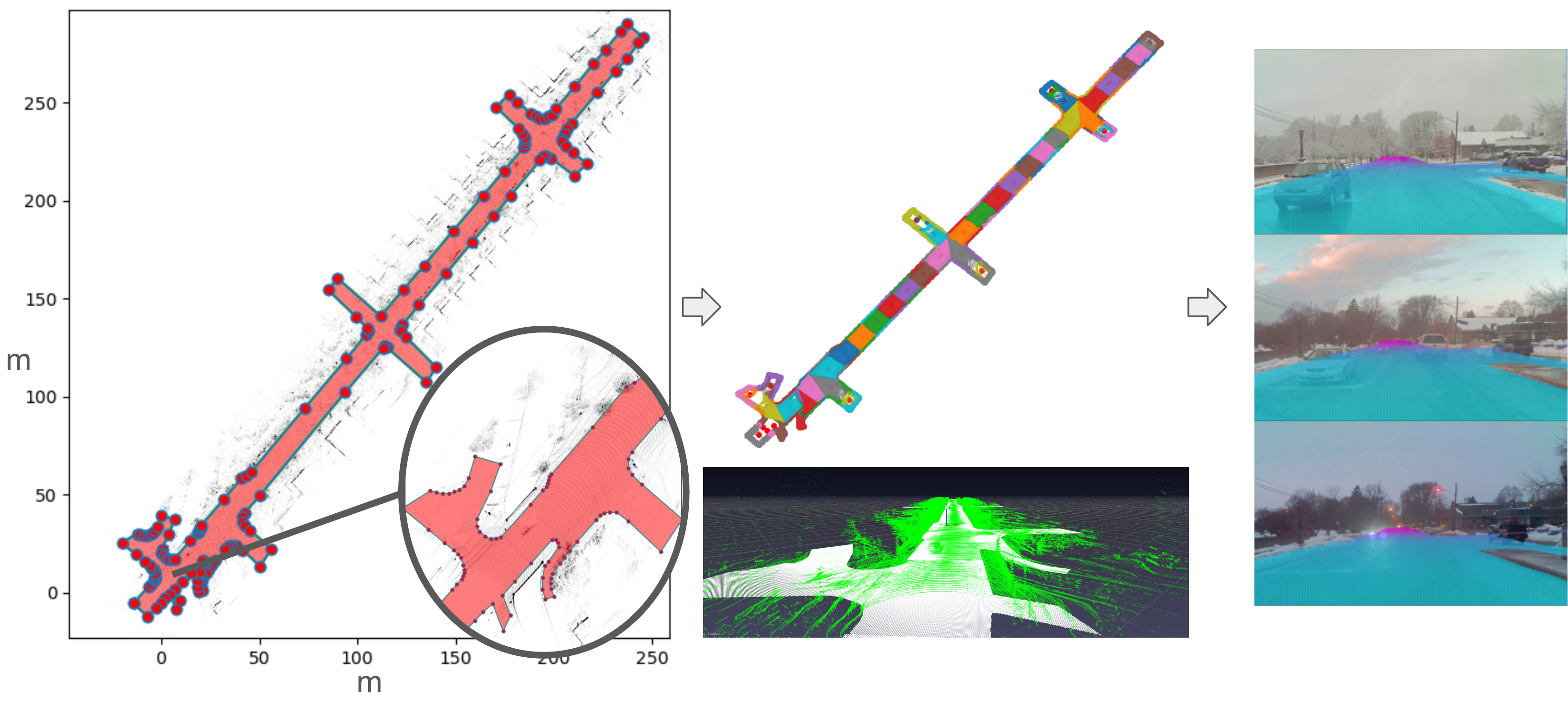}
  \caption{\small A built pointcloud is projected to BEV and the road is annotated using polygon labeling tool (left). The polygon is then divided into smaller 150 m$^2$ polygons and ground planes are estimated. Each color in the polygon represents a subdivision (middle). Finally ground planes are projected onto the image yielding amodal road mask with depth. Purple means farther distance (right)}
  \label{fig:labeling_road}
\end{figure}

\begin{figure}[ht]
  \centering 
  \includegraphics[width=\linewidth]{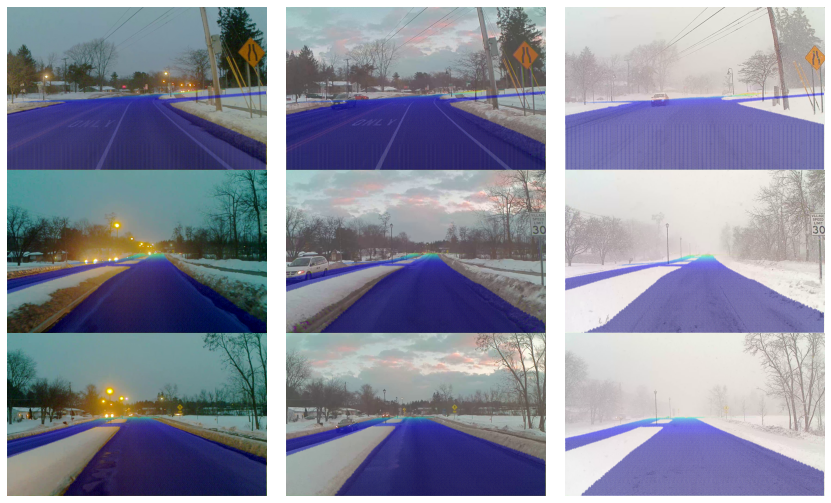}
  \caption{Generated ground truth depth masks for each pixel lying in the road area. Lighter blue indicates farther depth.}
 \label{fig:depth_masks}
\end{figure}

We show a labeling example in \autoref{fig:labeling_road} as explained in section \autoref{sec:labeling}.   Additional ground truth road depth masks are shown in Figure~\ref{fig:depth_masks}. 
\label{ssec:labeling}
\subsection{Sensor time synchronization details.}
\label{ssec:sync_details}
We use NVIDIA's recording tool to log the LiDAR at 10 Hz; cameras at 30 Hz. The Novatel GPS/INS data was logged at 100 Hz using the PC running ROS, time synchronized with the AGX through PTP 
from the Novatel custom firmware. The timing synchronization among cameras has been verified to average 60 $\mu$s. We select the OS2 as the reference to which all other sensors are matched. The  average time difference between this LiDAR and the cameras is 8.9 ms, with a worst case of 16.6 ms when there are no camera frames dropped. For the IMU/GPS, the average time difference is 3 ms; the worst case is 30 ms. The INS/GPS poses can be interpolated to the selected LiDAR time. For the VLPs, the worst case time difference is 35 ms. 
\subsection{Dataset Statistics}
\label{ssec:dataset statistics}
\begin{figure}[h!]
  \centering 
  \includegraphics[width=\linewidth, ]{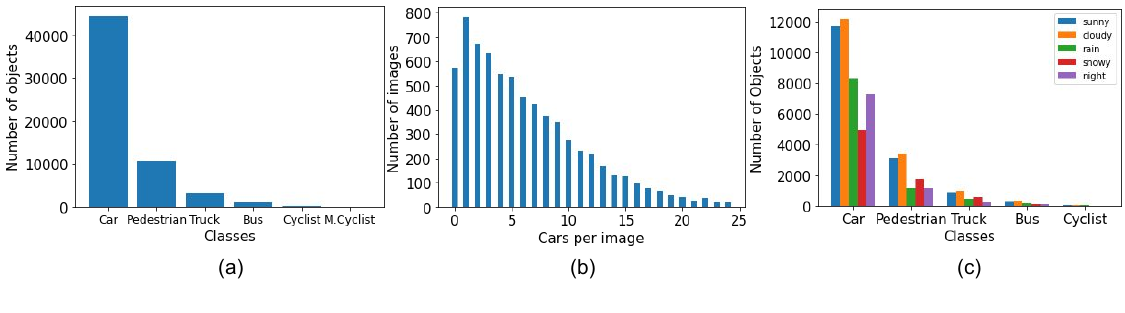}
  \caption{\small Statistics for a) overall object counts; b) cars per image; c) object distributions over weathers.}
\label{fig:statistics}
\end{figure}
We include statistics on the amodal object labels in Figure \ref{fig:statistics}.  Regarding the amodal road segmentation task the average number of close pixels per image are 767,759 while far are 38,552, which corresponds to 33.0\% and 1.7\% of the image, respectively. 
\section{Baseline Algorithm Details}
\label{ssec:baselines}
\subsection{Dual Attention network details}
\begin{figure}[h]
\minipage{0.5\textwidth}
\centering
  \includegraphics[height=0.35\linewidth]{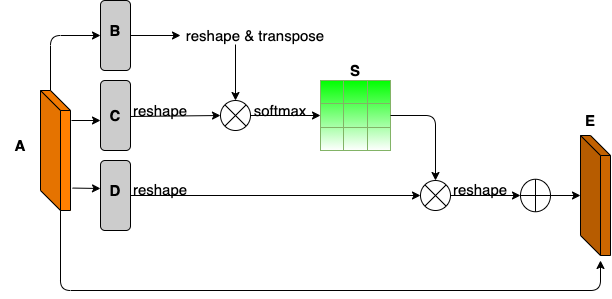}
\endminipage\hfill
\minipage{0.5\textwidth}
\centering
  \includegraphics[height=0.28\linewidth]{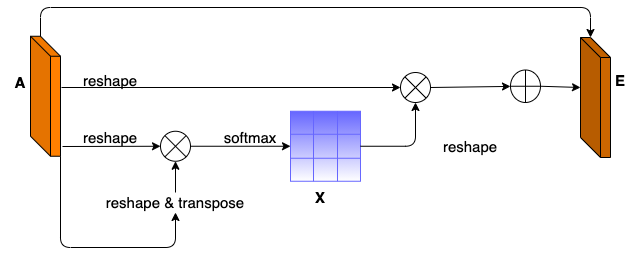}
\endminipage
\caption{The image here shows the PAM module (top) and the CAM module (bottom) proposed by~\cite{fu2019dual}}
\label{pam_cam}
\end{figure}

As discussed in \autoref{sec:baselines}, we add a positional attention module (PAM) and channel attention module (CAM) to our baseline. In Figure~\ref{pam_cam} we show diagrams for these two modules. 
\subsection{Pooling operation}
\begin{figure}[h]
  \centering 
  \includegraphics[width=\linewidth]{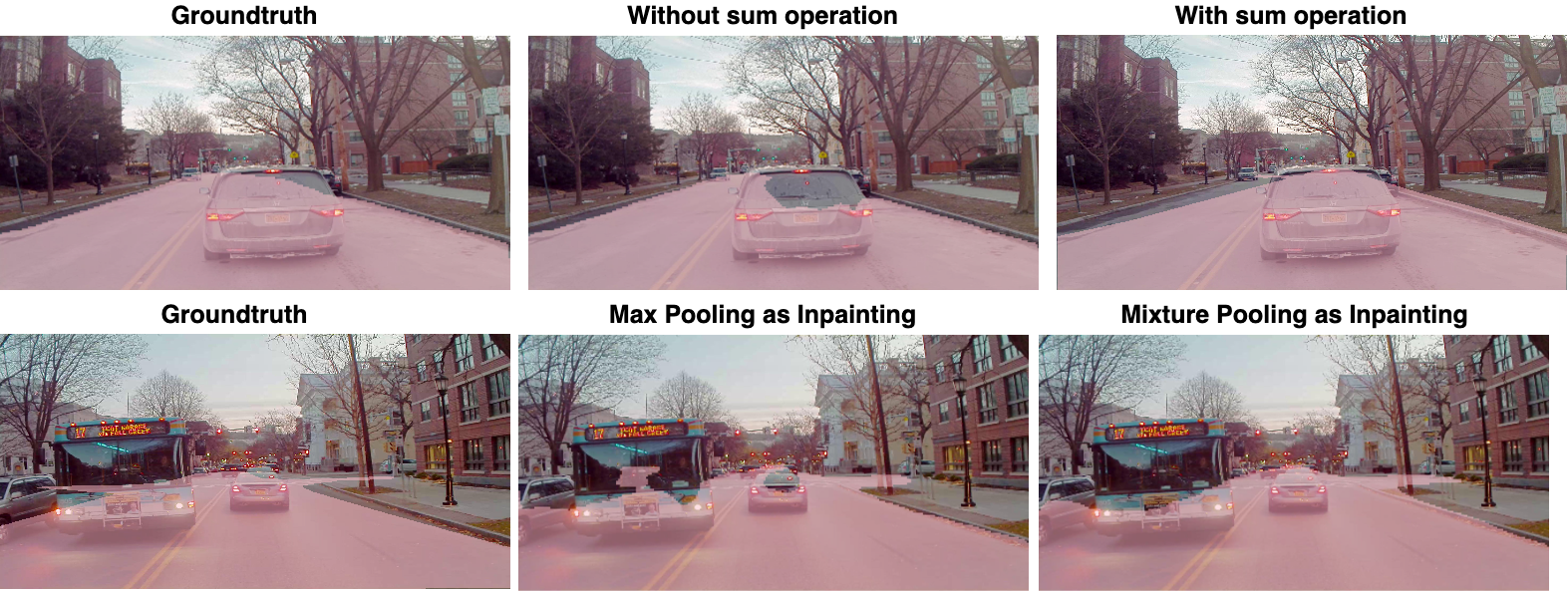}
  \caption{\small The top row shows our proposed baseline prediction results with Max Pooling as Inpainting and Mixture Pooling as Inpainting respectively. The bottom row shows our proposed baseline prediction results with and without sum operation in the road segmentation branch respectively.}
\label{fig:sum_and_pool}
\end{figure}

The detailed algorithm for Mixture Pooling as Inpainting is shown in Algorithm \ref{alg:mix}, mentioned in \autoref{sec:baselines}. This is a modification from Max Pooling as Inpainting, we refer the reader to \cite{lu2020semantic} for that algorithm. Qualitative results comparing Max Pooling vs Mixture pooling are demonstrated in Figure \ref{fig:sum_and_pool}.

    \begin{algorithm}[h]
        \caption{Mixture Pooling as Inpainting}
        \label{alg:mix}
        \begin{algorithmic}
        \Require $F_{raw}$ - intermediate feature map; $M$ binary foreground mask with 1 being background
        \Ensure $F_{inpainted}$ is inpainted feature map
        \State $M=$ max-pooling$(M)$
        \State $F_{background}=F_{raw}\times M$ 
        \State $F_{patch}=$ zero tensor with size same as $F_{background}$
        \State $M_{old}=M$
        \Do
        \State \textbf{$F_{background}=0.5*$ max-pooling $(F_{background})$ $+0.5*$ avg-pooling $(F_{background})$}
        \State $M_{new}=$ max-pooling $(M_{old})$
        \State $F_{patch}+=(M_{new}-M_{old})*F_{background}$
        \State $M_{old}=M_{new}$ 
        \doWhile{0 still exists in $M_{old}$} 
        \State $F_{inpainted}=F_{raw}*M+F_{patch}$
        \end{algorithmic}
    \end{algorithm}

\section{Training and inference code: amodal road}
\label{ssec:training_code}
Training and inference code is here: {\footnotesize{\url{https://github.com/coolgrasshopper/amodal_road_segmentation}}}
\\
\section{Training details \& visualization results}
\label{ssec:training_details}
\subsection{Hyperparameters \& Error analysis}
The detailed hyperparameters for our baseline network discussed in \autoref{sec:baselines} are demonstrated in our released source code. In general, we train the network using $240$ epochs with initial learning rate $0.3$ and weight decay $1e-04$ through the training process. Additionally, we set different seeds three times and record the error bars for our proposed baseline in Table~\ref{tbl:ablation}. Also, the standard deviation for the three conducted experiments are $0.042\%$ and $0.041\%$ for the \textit{far} and \textit{close} IOU respectively. For  both the \textit{close} and \textit{far} IOU, the errors and standard deviations are within $0.1\%$ using our proposed baseline. This shows that our experiments' results are reproducible.
\subsection{Qualitative performance evaluation of SFI}
\begin{figure}[h]
  \centering 
  \includegraphics[width=\linewidth]{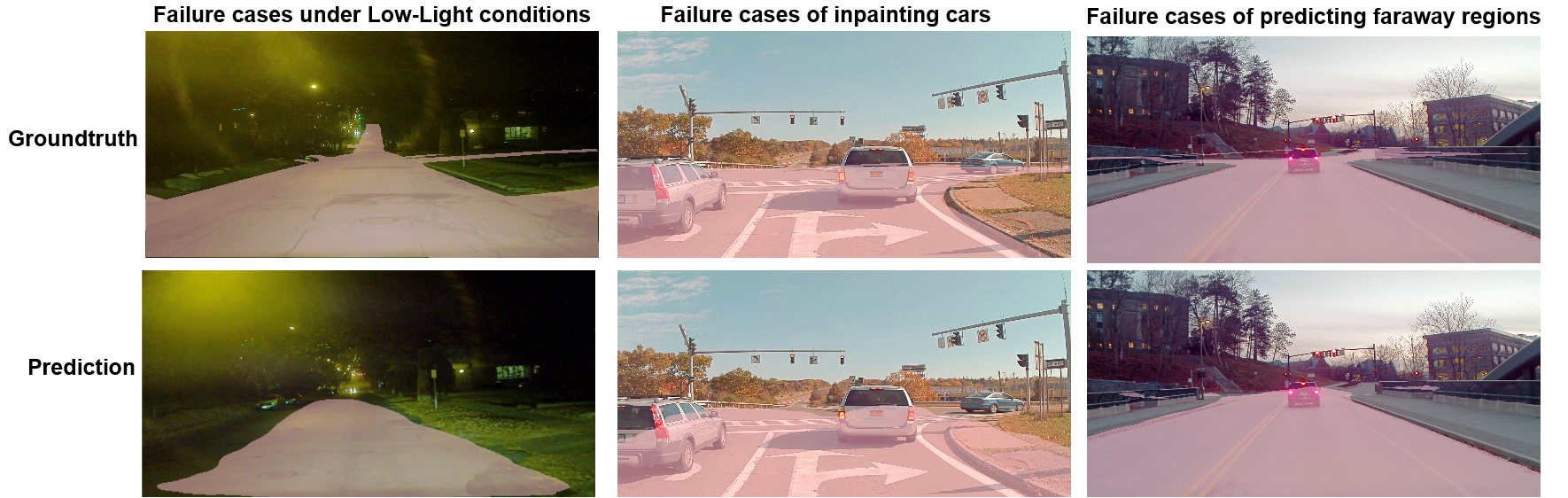}
  \caption{\small Visualized failure cases of the Semantic Foreground Inpainting network. Ground-truth overlay results are shown in the top row and network prediction results are in the bottom row.}
\label{fig:bad}
\end{figure}

The qualitative evaluation of SFI is shown in Figure~\ref{fig:bad} demonstrating the failure cases of SFI under challenging scenarios (\textit{e.g.} cluttered scenes, night and faraway regions).
\subsection{\textit{Far} \& \textit{Close} IOU calculation}
\label{sec:calc}
To calculate the \textit{far} and \textit{close} IOU, for each image $k$, we group each pixel $i$ based on its the depth $d_{ki}$. The set which contains all pixels with $d_{ki}<30$ is denoted as $DC_{k}$. The set that contains all pixels with $d_{ki}\geq 30$ is denoted as $DF_{k}$. After that, for the test dataset  that contains $N$ total test images, the \textit{close} IOU is calculated as:
\begin{equation}
    \frac{\sum_{k=0}^{N} I_{DC_{kg}}\cap I_{DC_{kp}}}{\sum_{k=0}^{N} I_{DC_{kg}}\cup I_{DC_{kp}}}
\end{equation}
where $I_{DC_{kg}}$ demonstrates for pixels with depth $d_{ki}<30$, the ground-truth binary mask for Amodal road segmentation at image $k$ and $I_{DC_{kp}}$ indicates the Amodal road segmentation prediction at image $k$ for \textit{close} pixels. 

Similarly, the \textit{far} IOU is calcuated as:
\begin{equation}
    \frac{\sum_{k=0}^{N} I_{DF_{kg}}\cap I_{DF_{kp}}}{\sum_{k=0}^{N} I_{DF_{kg}}\cup I_{DF_{kp}}}
\end{equation}
where $I_{DF_{kg}}$ demonstrates for pixels with depth $d_{ki} \geq 30$, the ground-truth binary mask for Amodal road segmentation at image $k$ and $I_{DF_{kp}}$ indicates the Amodal road segmentation prediction at image $k$ for \textit{far} pixels. 

We also attached the evaluation code in our anonymously released code: {\footnotesize{\url{https://github.com/coolgrasshopper/amodal_road_segmentation}}}
\subsection{Split by location results}
To investigate the effects of training on different road types and surrounding environments, we split the dataset into five different areas: urban, highway, rural, campus, downtown. Then, we train \textbf{our} proposed baseline using images collected at each location respectively. For each trained model, we test the model's performance using the dataset collected under the other four location types. The detailed performance is illustrated in Table~\ref{tbl:location}.
We observe that many models seem to drop in performance in the campus and urban environment. A potential reason for this is that urban areas contain more curved roads, and higher levels of occlusions and less visibility due to nearby buildings and vehicles, while campus contains some complex road structures with intersections and road islands (islets).

\begin{table}[h]
\footnotesize
\renewcommand{\arraystretch}{1.00}
\centering
\caption{\small \textbf{Results (IOU for road) on model performance among five different locations.}  On each entry (\emph{row}, \emph{column}), we report training on \emph{row} and testing on \emph{column} using our proposed baseline.\\ \label{tbl:location}}

\begin{tabular}[h]{r|ccccc}
    \toprule
    Far IOU & \multicolumn{1}{c}{Urban} & \multicolumn{1}{c}{Downtown} & \multicolumn{1}{c}{Highway} & \multicolumn{1}{c}{Rural} & \multicolumn{1}{c}{Campus} \\ \midrule
    Urban & N/A & 46.93 & 38.57 & 43.40 & 35.92 \\
    Downtown   &51.43& N/A & 51.93 & 55.81 & 41.55\\
    Highway   & 36.76& 46.11& N/A & 57.35 & 39.19 \\
    Rural   & 43.64& 47.18 & 53.92& N/A & 39.73 \\
    Campus  & 42.53&50.55& 53.63& 57.16 & N/A\\\midrule
   Close IOU & \multicolumn{1}{c}{Urban} & \multicolumn{1}{c}{Downtown} & \multicolumn{1}{c}{Highway} & \multicolumn{1}{c}{Rural} & \multicolumn{1}{c}{Campus} \\ \midrule
    Urban & N/A & 94.04 & 88.11 & 87.54 & 92.16 \\
    Downtown   &90.19 & N/A & 94.17 & 88.46 & 93.79\\
    Highway   & 86.54& 92.90& N/A & 88.27 & 91.36 \\
    Rural   & 92.89& 93.40 & 91.95& N/A & 91.17 \\
    Campus  & 92.95 &94.32& 94.21& 89.94 & N/A\\\bottomrule
    
\end{tabular}
\end{table}
\subsection{Ablation study of our proposed baseline}
We conduct an ablation study of our proposed model by removing the added sum operation in the road segmentation branch and not modifying the original Max pooling as Inpainting operation (\textit{i.e.}, removing the mixture pooling operation). That is, we remove the sum operation but keep Mixture pooling as Inpainting (third row of Table~\ref{tbl:ablation}).  We also keep the sum operation but use Max pooling as in painting (fourth row of Table~\ref{tbl:ablation}). Finally we also add the feature map from the channel attention module (fifth row of Table~\ref{tbl:ablation}). This channel attention module is to stress the inter-dependencies of feature maps. As each feature map can be regarded as a class-specific response, channel attention can also be interpreted as emphasizing the inter-dependencies of different (foreground) classes. For road segmentation, since we only have one semantic class, we remove the module to save computation in our proposed baseline. We test all trained models using the total collected test dataset and illustrate the \textit{close} and \textit{far} IOU. The results are shown Table~\ref{tbl:ablation}. We also demonstrate the \textit{close} and \textit{far} IOU performances of SFI and our proposed baselines in Table~\ref{tbl:ablation}.

\begin{table}[h]
\footnotesize
\renewcommand{\arraystretch}{1.00}
\centering
\caption{\small \textbf{Ablation study} that removes the sum operation (`w/o sum' row in the table) and the mixture pooling operation (`w/o mix pooling' row in the table) in the road segmentation branch respectively. We also include the \textit{close} and \textit{far} IOU for SFI and our proposed baselines in the table.\\ \label{tbl:ablation}}

\begin{tabular}{|l|l|l|}
\hline
Architectures   & \textit{Close} IOU & \textit{Far} IOU \\ \hline
SFI            &  91.55 &   52.16 \\ \hline
w/o sum         & 93.19     & 55.06   \\ \hline
w/o mix pooling & 93.08     & 54.77   \\ \hline
w/ $map_{CF}$ & 93.58 & 57.06 \\ \hline
Ours            & 93.29 ($\pm$ 0.072)    & 56.67 ($\pm$ 0.075)  \\ \hline

\end{tabular}
\end{table}

\subsection{Qualitative experiment results}

\begin{figure}[h]
\vspace{-2cm}
  \centering 
  \includegraphics[width=0.9\linewidth]{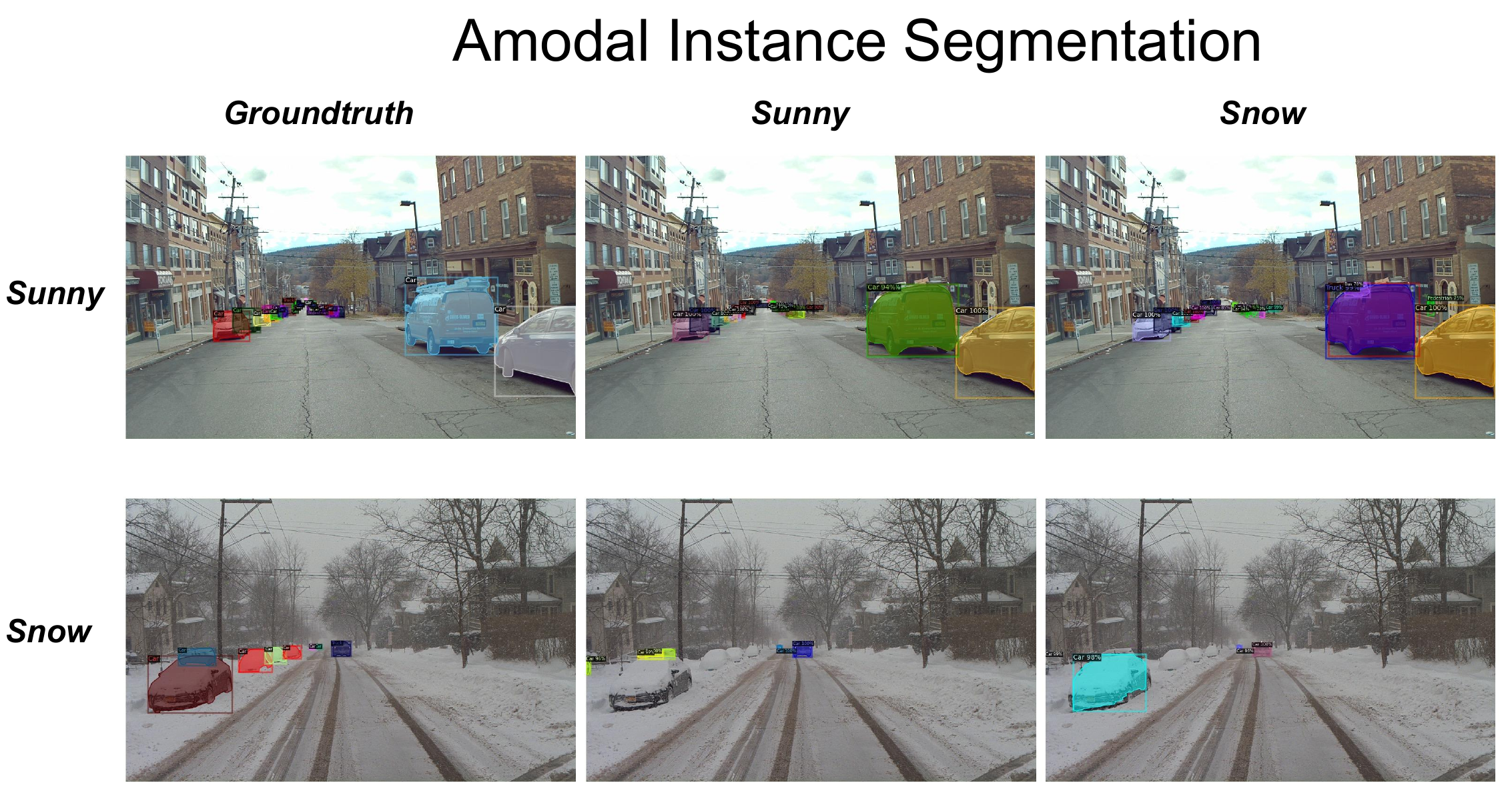}
  \caption{\small MaskRCNN model trained on the sunny (second column) and snowy (third column) datasets. The groundtruth is shown in the first column. The row indicates the condition being tested on (\ie sunny first row, snowy second row).  }
\label{fig:amodal_instance}
\end{figure}

\begin{figure}[h]
\vspace{-1cm}
  \centering 
  \includegraphics[width=\linewidth]{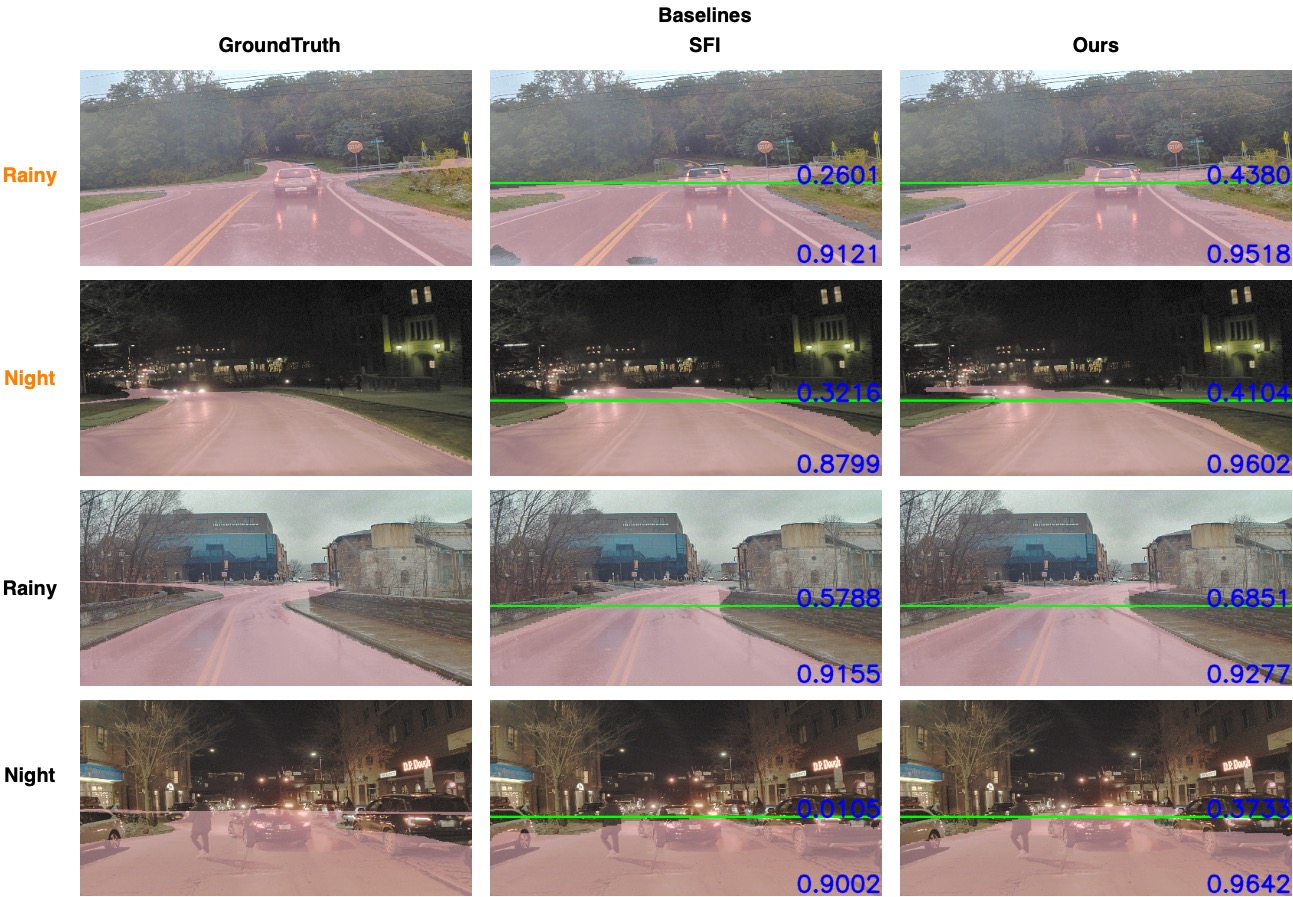}
  \caption{\small  Road inference for the two baselines. Orange indicates models trained on rainy with the first row testing on rainy and second row on night. Black indicates training on night with the third row testing on rainy and the fourth row testing on night. Above green line (30m) is the \textit{far} IOU and below it is \textit{close} IOU.}
\label{fig:res5} 

\end{figure}

\begin{figure}[h]
\vspace{-1cm}
  \centering 
  \includegraphics[width=\linewidth]{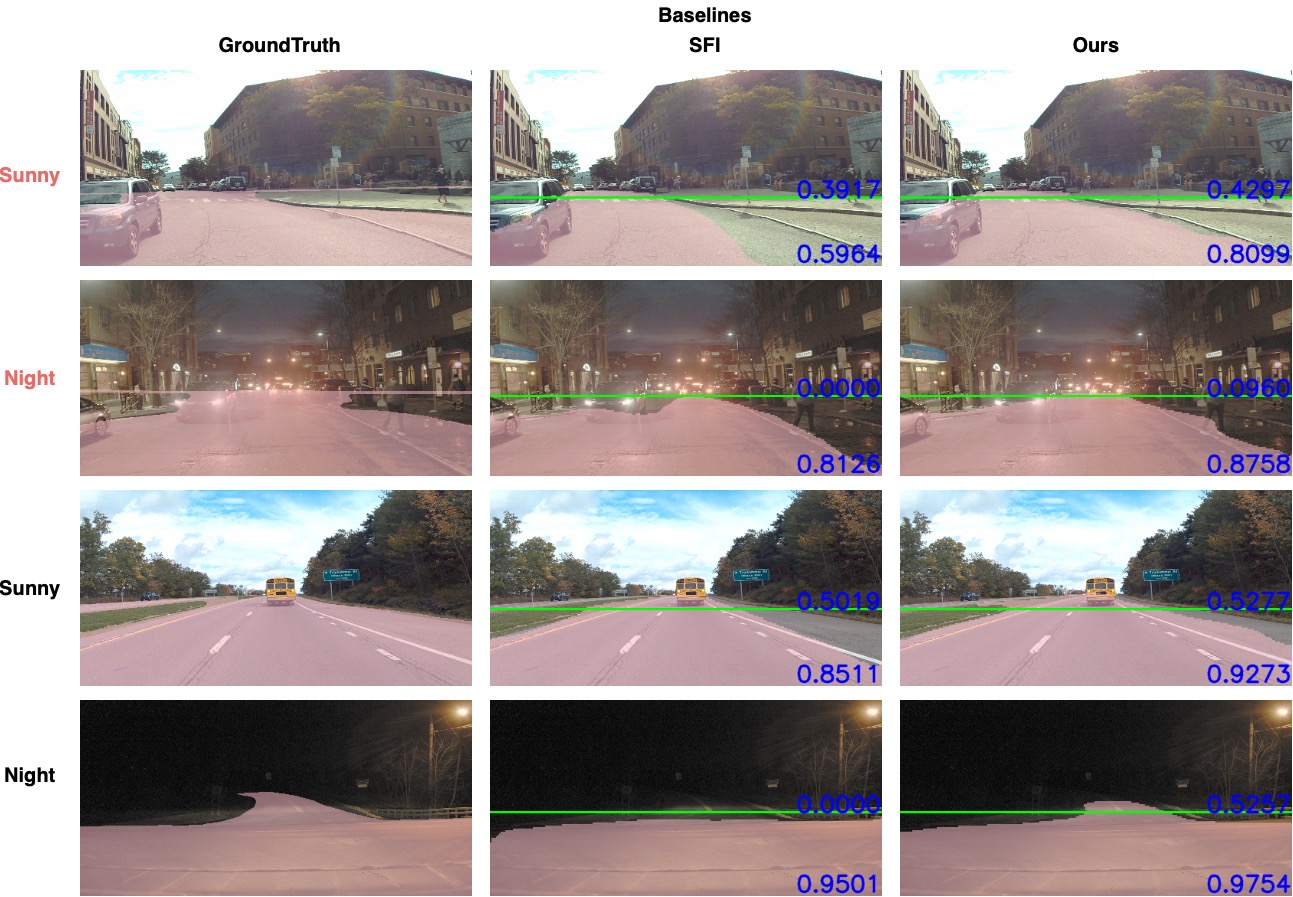}
  \caption{\small  Road inference for two baselines. Red indicates model trained on Sunny with the first row is testing on sunny and second row on night. Black indicates training on night with the third row testing on sunny and the fourth row testing on night. Above green line (30m) is the \textit{far} IOU and below it is \textit{close} IOU.}
\label{fig:res1}
\end{figure}

\begin{figure}[h]
\vspace{-2cm}
  \centering 
  \includegraphics[width=\linewidth]{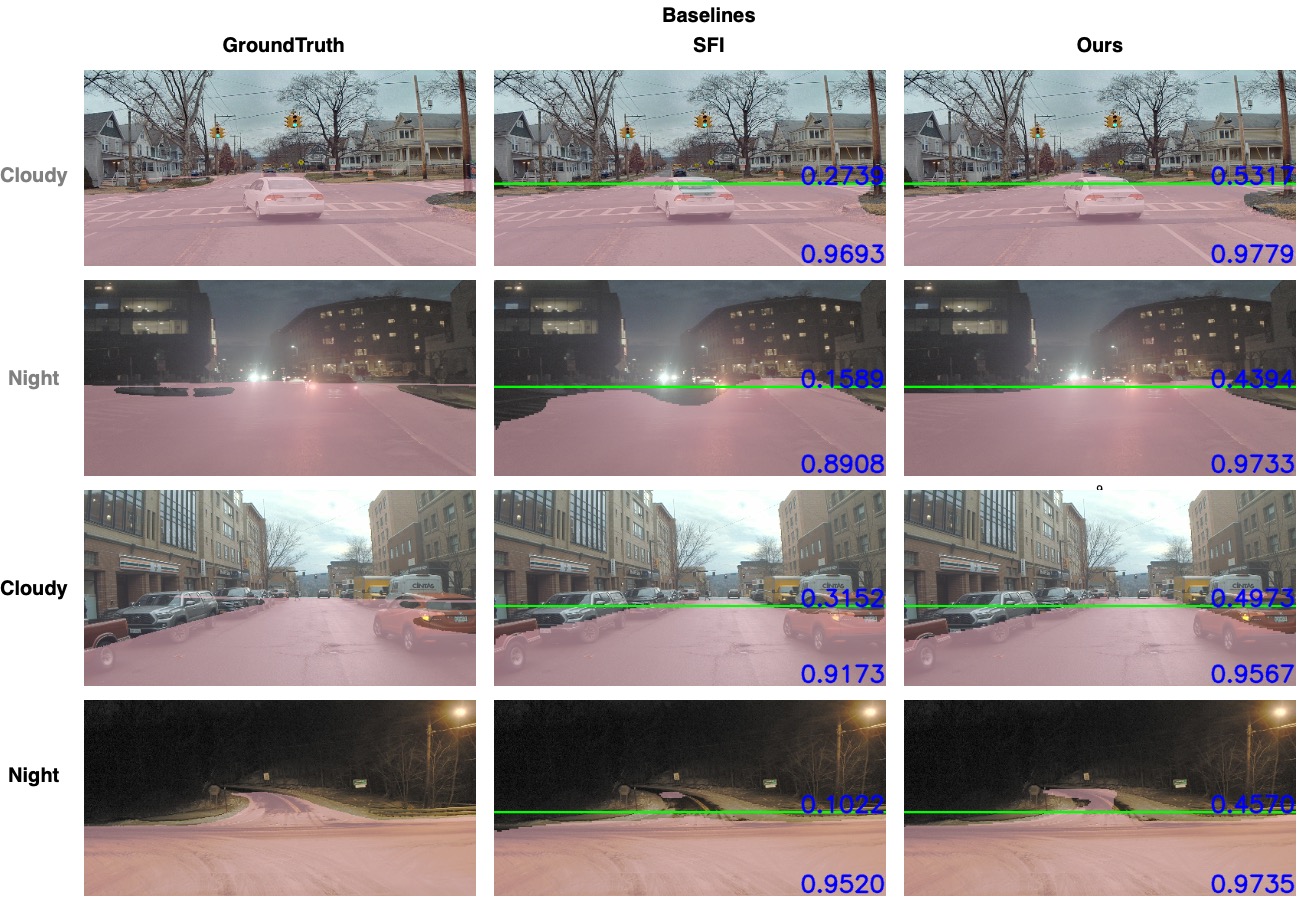}
  \caption{\small Road inference for two baselines. Grey indicates model trained on cloudy with the first row testing on cloudy and second row on night. Black indicates training on night with the third row testing on cloudy and the fourth row testing on night. Above green line (30m) is the \textit{far} IOU and below it is \textit{close} IOU.}

\label{fig:res2}
\end{figure}

\begin{figure}[h]
  \vspace{-0.5cm}
  \centering 
  \includegraphics[width=0.9\linewidth]{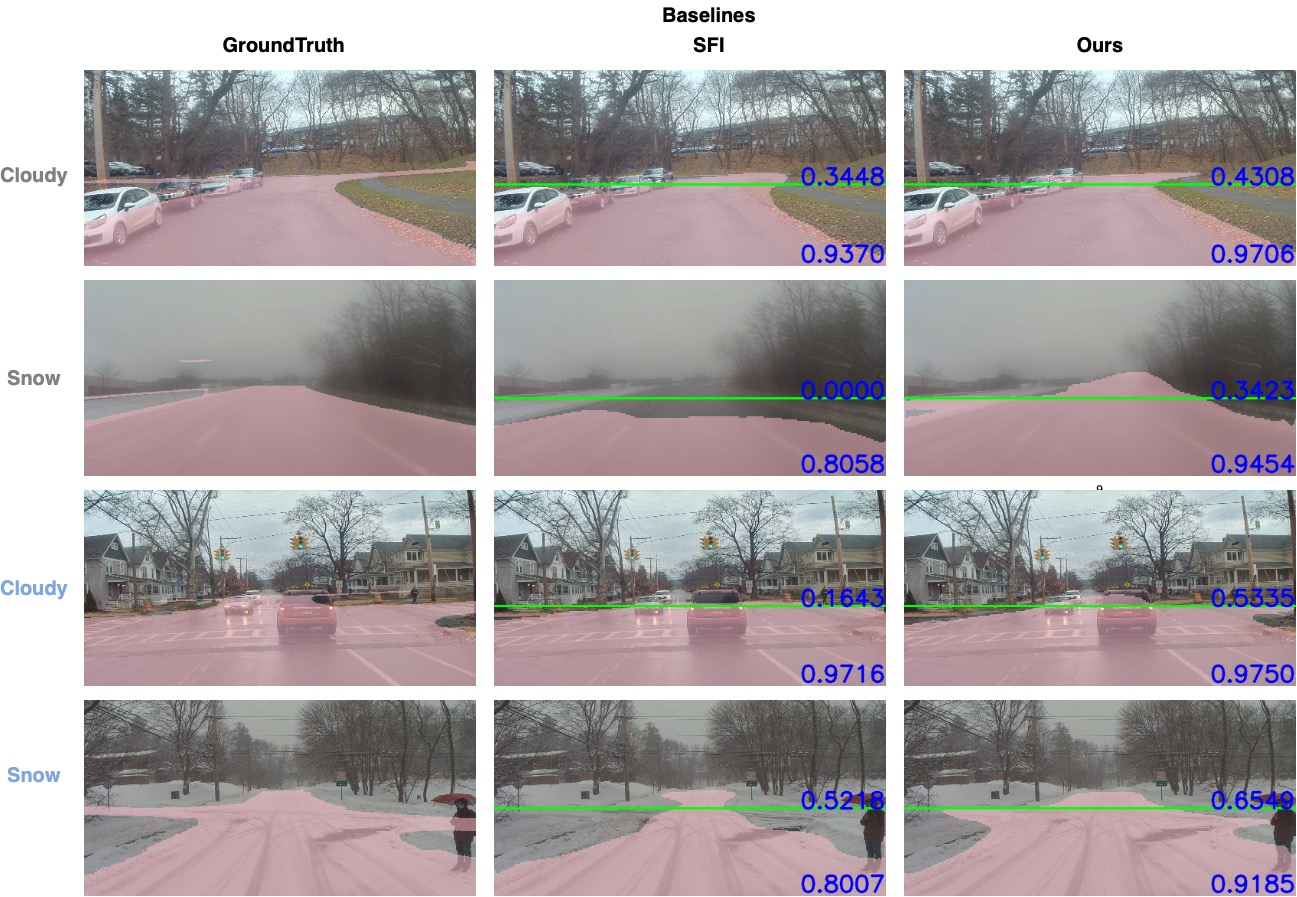}
  \caption{\small Road inference for two baselines. Grey indicates model trained on cloudy with the first row testing on cloudy and second row on snow. Blue indicates training on snow with the third row testing on cloudy and the fourth row testing on snow. Above green line (30m) is the \textit{far} IOU and below it is \textit{close} IOU.}
\label{fig:res3}

\end{figure}

\begin{figure}[h]
\vspace{-0.5cm}
  \centering 
  \includegraphics[width=0.9\linewidth]{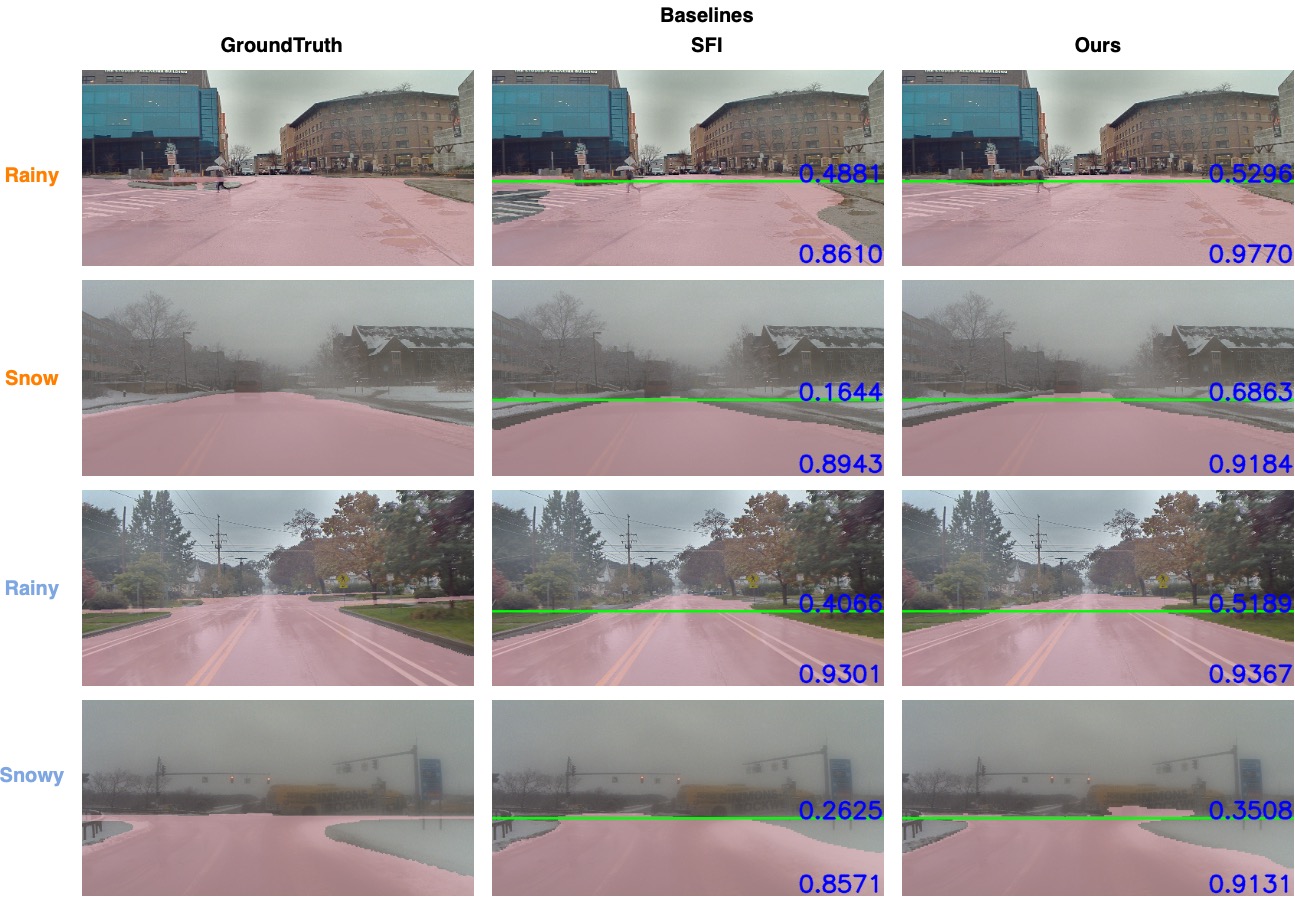}
   \caption{\small Road inference for two baselines. Orange indicates model trained on rainy with the first row testing on rainy and second row on snow. Blue indicates training on snow with the third row testing on rainy and the fourth row testing on snow. Above green line (30m) is the \textit{far} IOU and below it is \textit{close} IOU.}
\label{fig:res4}
\end{figure}

Finally, we demonstrate the visualization results of amodal instance segmentation in Figure~\ref{fig:amodal_instance}. From the visualization results, we find that cars and pedestrians are predicted more accurately under sunny situation than snowy and night situations. This underscores visual challenges for amodal scene reasoning under more adverse conditions. Qualitative results split by sunny, rainy, cloudy, night and snowy conditions are shown in Figure~\ref{fig:res5} to Figure~\ref{fig:res4} for amodal road segmentation. The green line demonstrates the \textbf{closest} horizontal line (height) to the bottom of the image which has depth $d_{h} \geq 30$. See Section~\ref{sec:calc} for detailed calculation).




\end{document}